\documentclass[pdflatex,sn-apa]{sn-jnl}
\usepackage{graphicx}%
\usepackage{multirow}%
\usepackage{amsmath,amssymb,amsfonts}%
\usepackage{amsthm}%
\usepackage{mathrsfs}%
\usepackage[title]{appendix}%
\usepackage{xcolor}%
\usepackage{textcomp}%
\usepackage{manyfoot}%
\usepackage{booktabs}%
\usepackage{algorithm}%
\usepackage{algorithmicx}%
\usepackage{algpseudocode}%
\usepackage{listings}%
\usepackage{makecell}

\usepackage{tikz}
\usetikzlibrary{positioning, calc, shapes.geometric, arrows.meta, matrix, fit}
\usepackage{float}
\usepackage{xspace}
\usepackage{subcaption}
\usepackage{hyperref}
\usepackage{lmodern}
\usepackage{pgfplotstable}
\pgfplotsset{compat=1.18}
\floatstyle{plain}
\newfloat{listing}{htbp}{lop}
\floatname{listing}{Listing}

\definecolor{mygreen}{rgb}{0,0.6,0}
\definecolor{mygray}{rgb}{0.5,0.5,0.5}
\definecolor{mymauve}{rgb}{0.58,0,0.82}

\definecolor{codegreen}{rgb}{0,0.6,0}
\definecolor{codegray}{rgb}{0.5,0.5,0.5}
\definecolor{codepurple}{rgb}{0.58,0,0.82}
\definecolor{backcolour}{rgb}{0.95,0.95,0.92}
\definecolor{valuecolor}{RGB}{128, 0, 128}

\lstdefinelanguage{YAML}{
	keywords={true,false,null,y,n},
	keywordstyle=\color{blue},
	comment=[l]{\#}, 
	commentstyle=\color{mygreen},
	stringstyle=\color{codepurple},
	identifierstyle=\color{brown},
	sensitive=false,
	alsoletter={:},
	morekeywords=[2]{1,2,3,4,5,6,7,8,9,0},
	keywordstyle=[2]\color{valuecolor},
	morekeywords=[3]{True,False},
	keywordstyle=[3]\color{valuecolor},
	breaklines=true,
	breakatwhitespace=true,
	tabsize=3,
	backgroundcolor=\color{backcolour},
	basicstyle=\footnotesize\ttfamily,
	showstringspaces=false
}

\newcommand{\YOLOvEight}{\textit{YOLOv8 }}
\newcommand{\YOLO}{\textit{YOLO}}

\newcommand{\orcidlogo}{\includegraphics[width=1em]{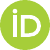}} 
\newcommand{\orcid}[1]{\href{https://orcid.org/#1}{\mbox{\orcidlogo}}}

\title{hYOLO Model: Enhancing Object Classification with Hierarchical Context in \YOLOvEight}

\date{} 

\begin{document}
	\maketitle
	
	\begin{center}
		\textbf{
	Veska Tsenkova\textsuperscript{1*}\orcid{0009-0009-2827-4146}, 
	Peter Stanchev\textsuperscript{1\textdagger}\orcid{0009-0000-7059-3778}, \\
	Daniel Petrov\textsuperscript{1\textdagger}\orcid{0009-0000-8599-8612}, 
	Deyan Lazarov\textsuperscript{1\textdagger}\orcid{0009-0004-3431-1680}
}		

		\vspace{1em}

\textsuperscript{1}DS and AI Solutions, Soft2RUN, Sofia, 1000, Bulgaria

\vspace{1em}

		\textbf{Corresponding author}: \href{mailto:vtsenkova@soft2run.com}{vtsenkova@soft2run.com} \\
		
		\vspace{1em}

		\textbf{Contributing authors}: 
		\href{mailto:pstanchev@soft2run.com}{pstanchev@soft2run.com}; 
		\href{mailto:hyolopermission@soft2run.com}{hyolopermission@soft2run.com}; 
		\href{mailto:dlazarov@soft2run.com}{dlazarov@soft2run.com}
		
		\textsuperscript{\textdagger}These authors contributed equally to this work.
		
	\end{center}

\vspace{2em}

\maketitle
\begin{abstract}
\noindent\textbf{Abstract:} Current convolution neural network (CNN) classification methods are predominantly focused on flat classification which aims solely to identify a specified object within an image. However, real-world objects often possess a natural hierarchical organization that can significantly help classification tasks. Capturing the presence of relations between objects enables better contextual understanding as well as control over the severity of mistakes. Considering these aspects, this paper proposes an end-to-end hierarchical model for image detection and classification built upon the \YOLO \xspace model family. A novel hierarchical architecture, a modified loss function, and a performance metric tailored to the hierarchical nature of the model are introduced. The proposed model is trained and evaluated on two different hierarchical categorizations of the same dataset: a systematic categorization that disregards visual similarities between objects and a categorization accounting for common visual characteristics across classes. The results illustrate how the suggested methodology addresses the inherent hierarchical structure present in real-world objects, which conventional flat classification algorithms often overlook.
\end{abstract}

\vspace{1em}

\noindent\textbf{Keywords:} Computer vision, Hierarchical classification, Loss function, \YOLO \xspace model

\section{Introduction}\label{intro}
Many real-world classification problems are structured hierarchically, with target classes organized into multiple levels of abstraction. In such cases, a hierarchical classification model aims not only to predict the correct class but also to do so within a taxonomy that reflects the relationships among classes. For example, in biological datasets, organisms are classified into a hierarchical structure consisting of kingdom, phylum, class, order, family, genus, and species. A \textbf{flat} classification model directly classifies all the species without considering the hierarchical relationships, bypassing the broader categories such as kingdom, phylum, or class. In contrast, a \textbf{hierarchical} classification model begins by predicting the broadest category, the kingdom, and progressively refines its predictions as it moves down the taxonomy, ultimately reaching the most specific level, the species. This approach ensures that predictions are made within the context of related classes, thereby capturing the hierarchical structure and enabling the model to better manage complex relationships between different levels. In addition, hierarchical classification models offer several advantages over flat models, particularly in domains like computer vision, where large number of classes and hierarchical levels need to be processed:

\textbf{Error control and More Efficient Learning}: 

By teaching the network to maintain predicted classes within the same hierarchical category, hierarchical models not only learn how to accurately classify objects but also how to make errors that are less harmful. This feature is particularly valuable in high-risk domains such as medical image diagnosis, where the consequences of misclassification can be significant. Rather than treating each diagnosis as an independent class, the model may first classify a skin lesion into broad risk groups such as ``benign,'' ``pre-malignant,'' or ``malignant,'' before refining the prediction to a specific condition. This structure reduces the likelihood of critical errors, such as misclassifying a malignant tumor as benign, by keeping misclassifications within the same risk category. A representative example is the work by \citet{DBLP:conf/miccai/YuNGJCZBMWG22}, who proposed a class-hierarchy regularized hyperbolic embedding model for skin lesion recognition. Their results show that incorporating hierarchical structure improves both accuracy and diagnostic safety by aligning misclassifications with clinically similar categories.

In addition, instead of learning all classes simultaneously, their hierarchical model progresses through multiple levels of abstraction, beginning with broad, coarse-grained categories and incrementally refining its predictions toward more specific target classes. The model initially groups lesions into ``benign,'' ``pre-malignant,'' or ``malignant'' categories, and then further differentiates within each group (e.g., distinguishing between ``basal cell carcinoma'' and ``squamous cell carcinoma'' within the malignant category). This stepwise, sequential learning process enhances the model's convergence rate and improves generalization, as it focuses on high-level features before addressing more complex details.

\textbf{Reduced Complexity}: In complex classification tasks with a large number of classes, flat classification models may struggle to learn simultaneously discriminative features for each class. Hierarchical models manage complexity by breaking down classification tasks into manageable subproblems. The model initially deals with simpler, broader classes, and gradually adds more detailed predictions as the hierarchy deepens. This reduces the difficulty of learning discriminative features for every class at once, making the task more manageable. For example, a hierarchical classification model diagnosing lung conditions (\citet{DBLP:conf/miccai/YangGKNSXC20}) may first classify images into broader categories, such as ``healthy'' or ``abnormal.'' Within the ``abnormal'' category, the model could further differentiate between conditions like ``pneumonia,'' ``tuberculosis,'' or ``lung cancer.'' This structured approach allows the model to focus on learning general patterns for healthy vs. abnormal tissue, and then focus on the more challenging task of distinguishing between various diseases within the abnormal group.

\textbf{Improved Interpretability and Scalability}: 

Hierarchical models naturally capture semantic relationships between classes, enhancing both interpretability and scalability: two critical features for domains such as healthcare or inventory management. In terms of interpretability, these models enable users to trace predictions through successive layers of abstraction, offering insight into how decisions are made. For example, in chest X-ray analysis (\citet{DBLP:journals/mia/ChenMXHH20}), a model may move from broad labels like ``normal'' or ``abnormal'' to increasingly specific diagnoses (e.g., pneumonia, lung cancer, or COPD), aligning with the clinician’s diagnostic reasoning. This multi-level hierarchy allows radiologists to follow a clear progression from a general abnormality to a very specific diagnosis, making it easier to understand the model’s results and assess its reliability. From a scalability perspective, the same hierarchical structure reduces the complexity of large-scale classification tasks by decomposing them into smaller, more manageable subproblems. Instead of learning discriminative features for thousands of flat classes simultaneously, the model progressively narrows down predictions by traversing the hierarchy, improving computational efficiency and making it feasible to handle extensive class taxonomies without degrading performance.

Despite the clear advantages of hierarchical classification, the majority of CNN-based classification methods continue to predominantly rely on flat models, where each class is treated independently, disregarding any relationships between them. In these models, the network must simultaneously differentiate between all possible classes, a process that not only amplifies task complexity but also demands substantial computational resources, especially when handling large number of classes. Furthermore, flat models fail to account for the varying severity of misclassifications within a hierarchical context. A primary reason for the continued preference for flat models is the scarcity of readily available hierarchical datasets. Constructing such datasets is challenging, requiring careful definition of meaningful hierarchies and precise labeling across multiple levels. Additionally, evaluating hierarchical models requires the development of hierarchical performance metrics and loss functions that accurately capture hierarchical class relationships. The lack of structured datasets, appropriate evaluation metrics, and specialized loss functions makes it difficult to train models that exploit class relationships, thus preserving the dominance of flat models in many domains. This paper addresses these limitations by introducing an end-to-end hierarchical classification model built upon the \YOLO \xspace model family. Experimental results highlight how our hierarchical approach overcomes the challenges inherent in flat models, particularly in the real-world task of grocery store item classification.

\section{Related Work}\label{related_work}
Incorporating hierarchical structures into deep learning tasks has proven to be a successful approach in various domains, including computer vision and text classification. Research efforts have largely focused on adapting neural network architectures, loss functions, and label representations to better capture hierarchical relationships between classes. 

\textbf{Architectural adaptations} commonly include designs that reflect hierarchical dependencies explicitly: adding multiple output layers corresponding to different levels of the class hierarchy, designing custom layers that encode hierarchical relationships, or integrating graph-based structures.  For example, \citet{DBLP:journals/corr/abs-1709-09890} introduced the Branch Convolutional Neural Network (\textbf{B-CNN}), which contains multiple branch networks along the main convolutional path, each corresponding to a different level in the class hierarchy. Building on this, \citet{DBLP:journals/remotesensing/TaoufiqNB20} replaced parallel branches with a coarse-to-fine classification strategy and introduced a multiplicative layer to explicitly model dependencies between coarse and fine predictions, resulting in fewer parameters and improved hierarchical consistency in their \textbf{HierarchyNet}. With \textbf{Tree-CNN} \citet{DBLP:journals/nn/RoyPR20} further advanced hierarchical modeling by organizing classifiers in a tree structure that can grow incrementally as new classes appear, enabling scalable learning without retraining from scratch. Another architecture-level innovation by \citet{DBLP:journals/corr/abs-2212-04161} proposed classifier-block-level hierarchies, reducing redundancy and memory consumption relative to traditional network-level hierarchical models. An alternative approach by \citet{DBLP:journals/apin/GrassaGL21} modified the output layers of a standard ResNet18, incorporating multiple linear layers for different hierarchy levels, and combined cross-entropy loss which aims to maximize inter-class variance,  with center loss (\cite{DBLP:conf/eccv/WenZL016}) to minimize intra-class variance. These architectural innovations demonstrate a range of methodologies for embedding hierarchical awareness directly into network design, improving classification accuracy, scalability, and adaptability to complex label taxonomies. Building on this foundation, our novel architecture integrates multiple hierarchical layers, performing convolutional operations at each level to effectively capture class dependencies and enhance hierarchical consistency throughout the model.

In parallel, \textbf{loss function engineering} has addressed the challenge of incorporating hierarchical information to improve model training. \citet{DBLP:conf/visapp/MullerS20} proposed a Hierarchical Loss for semantic segmentation that penalizes errors according to the semantic distance between predicted and true labels, encouraging semantically plausible mistakes when exact classification cannot be achieved. In a related approach, \citet{DBLP:conf/cvpr/BertinettoMTSL20} designed a Hierarchical Cross-Entropy Loss that increases penalties for misclassifications across distant hierarchy branches. \citet{DBLP:conf/wacv/Kobayashi21} proposed a hierarchy-aware training approach using soft hierarchical targets to share information among related classes, improving generalization especially for imbalanced datasets. \citet{DBLP:conf/ijcai/GoyalCG21} developed a Hierarchical Class-Based Curriculum Loss, guiding the model to learn coarse distinctions before fine-grained ones by using the hierarchy as a curriculum. These methods collectively illustrate diverse strategies—from pixel-level penalty adjustment (\citet{DBLP:conf/visapp/MullerS20}) and curriculum learning (\citet{DBLP:conf/ijcai/GoyalCG21}) to severity-weighted loss (\citet{DBLP:conf/cvpr/BertinettoMTSL20}) and soft target regularization (\citet{DBLP:conf/wacv/Kobayashi21}) — to integrate hierarchical structures into loss functions and model training.

Focusing specifically on \textbf{hierarchical object classification with \YOLO \xspace based models}, \citet{DBLP:conf/cvpr/RedmonF17} extended the original \YOLO \textit{v2} model to \YOLO \textit{9000}, to classify a significantly larger number of classes, up to $9000$, by introducing the WordTree hierarchical concept. To classify a specific object, the model calculates the conditional probability at each node level and then traverses the tree from the target node up to the root, multiplying these probabilities along the way to obtain the final class probability. While pioneering, \YOLO \textit{9000} has limited flexibility due to its reliance on a fixed, coarse-grained WordTree hierarchy, which may not align with domain-specific taxonomies. Additionally, it lacks hierarchy-aware loss functions and intermediate-level supervision, and is built on the outdated \YOLO \textit{v2} architecture, which falls short in accuracy and efficiency compared to modern models. As an application-specific extension, \citet{Kalhagen2022} proposed a \YOLO \xspace FISH hierarchical model to identify fish species in underwater video feeds and classify them in seven classes. This model adapts the \YOLO \textit{9000} framework by modifying the Non-Maximum Suppression (NMS) technique to remove redundant bounding boxes irrespective of object class. Furthermore, the \YOLO \textit{v3} detection layers were substituted with those from \YOLO \textit{9000}, enhancing the model's ability to detect hierarchies. 

The broader \YOLO \xspace framework has undergone rapid evolution in recent years. From \YOLO \textit{v3} through \YOLO \textit{v5}, improvements centered on multi-scale prediction, CSPNet-based architectures, and performance optimizations in both accuracy and inference speed. \YOLO \textit{v6} and \textit{v7} introduced dynamic label assignment strategies and more efficient training routines. \YOLO \textit{v8} brought architectural refinements such as decoupled head structures, anchor-free detection, and increased modularity (\citet{DBLP:journals/make/TervenER23}). Despite these advancements, hierarchical classification focused on comprehensive class taxonomies remains largely underdeveloped within the \YOLO \xspace family, with existing examples typically confined to narrow, domain-specific applications rather than forming a generalized framework. For example, \citet{usmani2025hierarchical} introduced Hierarchical \YOLO \xspace with Real-Time Text Recognition for UAE traffic signs, combining hierarchical detection with embedded text recognition to improve accuracy in complex environments. \citet{DBLP:journals/corr/abs-2407-17906} proposed a Hierarchical Object Detection and Recognition Model for practical plant disease diagnosis, using hierarchical structures to improve robustness and interpretability. Recent advancements have mostly concentrated on incorporating hierarchical context into feature extraction, rather than developing models that perform classification across hierarchical class structures. HCA-YOLO (\citet{DBLP:journals/corr/abs-2408-04804}) embedded a hierarchical coordinate attention mechanism within \YOLO \textit{v8}’s backbone to enhance multi-level feature representation. Similarly, HGO‑YOLO (\citet{DBLP:journals/corr/abs-2503-07371}) enhances \YOLOvEight by integrating HGNetv2 within the backbone to perform hierarchical feature extraction. Though promising, such methods focus on hierarchical features rather than hierarchical class labels. Our method addresses this gap by introducing a flexible, domain-adaptable hierarchical classification structure alongside an updated \YOLO \xspace backbone, to improve detection accuracy and adaptability across diverse application areas.

The \textbf{evaluation of hierarchical classification models} has also become an important area of research, with growing attention to metrics that account for the structure of class hierarchies. \citet{DBLP:conf/ai/KiritchenkoMNF06} introduced a measure, which considers distance and depth in the class hierarchy, crediting partially correct classifications and discriminating between different types of errors. In our implementation within \YOLO, we employed this measure to assess and compare the performance of different architectures of the hierarchy. Another hierarchical measure proposed by \citet{DBLP:journals/datamine/KosmopoulosPGPA15} evaluated pairs of predicted and true classes, assigning costs based on hierarchical distances. In addition, class relationships were represented as a network flow problem, minimizing classification error by pairing classes optimally.

\textbf{Recent research} in hierarchical computer vision has focused on developing scalable, modular models that effectively capture class dependencies across multiple levels of abstraction through varied core mechanisms. For instance,\citet{electronics12224646} introduce Hierarchical PPCA, which trains independent Probabilistic PCA models per class and clusters them into super-classes, significantly reducing classification complexity and speeding inference for large-scale datasets. In contrast, \citet{DBLP:conf/icann/MayoufS22} present a unified CNN architecture that simultaneously predicts labels at all hierarchy levels using Bayesian adjustments to encode class dependencies and a semantic loss to enforce hierarchical consistency. \cite{DBLP:journals/bspc/HuoSTWYLZL24} integrate CNN and Transformer branches within a multi-scale hierarchical framework, using an adaptive fusion module to combine local and global features, improving accuracy on medical image classification. Extending the use of attention mechanisms, a triplet attention-based model for robotic perception by \citet{DBLP:journals/sivp/BhayanaV24} introduces hierarchical supervision by jointly predicting object class and hierarchical position, thus enhancing performance on structured datasets. Despite differences in modeling choices, from probabilistic and CNN-based methods to attention and transformer fusion, all these approaches maintain hierarchical consistency and improve efficiency in  multi-level classification tasks.

\section{Methods}\label{methods}
By implementing a natural hierarchical object organization, the proposed end-to-end hierarchical model for image detection and classification, hYOLO, based on \YOLOvEight (\cite{ultralytics_yolo}), offers several innovative contributions.

Firstly, the hierarchical architecture of hYOLO captures the inter-class relationships between objects by organizing them into a meaningful hierarchical structure. This structure allows the model to understand the contextual relationships between different classes of objects, enabling more accurate and contextually relevant classifications. 

Secondly, the modified loss function is designed to penalize errors based on their severity within the hierarchy. A misclassification of a bottle of Reduced-Fat Milk (2\% fat) as Whole Milk (3\% fat) is penalized less than mistaking a bottle of Reduced-Fat Milk for wine. Thus, by incorporating this hierarchical penalty scheme into the loss function, the model learns to prioritize more critical distinctions, leading to improved overall performance.

In addition, a performance metric that reflects the hierarchical nature of the classification task is implemented. It provides a more nuanced evaluation of model performance by considering the hierarchical relationships between classes.

The implementation of the proposed model requires minor adjustments to the existing \YOLO \xspace framework, such as adapting the input label format to reflect hierarchical relationships and incorporating hierarchical performance metrics for each hierarchical level in the output file. Apart from these adjustments, the model training and evaluation procedures remain consistent with the original \YOLO \xspace setup. However, due to its novel architecture, hYOLO model must be trained independently; utilizing an already trained \YOLO \xspace model and retraining it is not feasible.

\subsection{Hierarchical Architectures}\label{hier_architectures}
A class taxonomy can be referred to as a hierarchy when its structure satisfies the following conditions:

\begin{enumerate}[parsep=1pt]
	\setlength{\itemsep}{0.5pt}
	\setlength{\parskip}{0pt}
	\setlength{\parsep}{0pt}
	\item It starts with a single root node or top-level class that contains all other classes within the taxonomy. From this root node, the hierarchy branches out into multiple levels, each level representing a subset of classes;
	\item Classes are arranged hierarchically, each class has one or more child classes that inherit properties or characteristics from their parent class. This establishes a hierarchical relationship between classes;
	\item The structure is acyclic, i.e., two classes cannot be each others ancestor, there are no loops in the hierarchy. This ensures that each class is uniquely positioned within the hierarchy;
	\item The structure is anti-reflexive, i.e., a class cannot be a parent of itself.
	\end {enumerate}
	
	The most popular hierarchical structures are Directed Acyclic Graphs (DAGs) which allow for multiple paths to the same node, and trees which ensure a unique path from the root to any specified node. For our hierarchy, we opted for a tree-based classifier due to its simplicity and ease of comprehension. 
	
\begin{figure}[htb!]
		\centering
		\fbox{\rule{0pt}{0.5in}
			\centering
			\captionsetup[subfigure]{position=above}
			
			\begin{subfigure}[b]{0.22\textwidth}
				\centering
				\caption{Local Classifier \\ per Node \textbf{LCN}}
				\raisebox{-0.5\height}{%
					\begin{tikzpicture}[node distance=0.3cm, scale=0.4]
						\node[scale=0.5, font=\Large] (R) at (0, 0) {R};
						\node[scale=0.5, font=\Large, below right=of R] (B) {B};
						\node[scale=0.5, font=\Large, below left=of R] (A) {A};
						\node[scale=0.5, font=\Large, below left=of A, yshift=-0.5cm, xshift=0.5cm] (C) {C};
						\node[scale=0.5, font=\Large, below right=of A, yshift=-0.5cm, xshift=-0.5cm] (D) {D};                
						\node[scale=0.5, font=\Large, below left=of B, yshift=-0.5cm, xshift=0.5cm] (E) {E};
						\node[scale=0.5, font=\Large, below right=of B, yshift=-0.5cm, xshift=-0.1cm] (F) {F};                       
						\draw[->] (R) -- (A);
						\draw[->] (R) -- (B);
						\draw[->] (A) -- (C);
						\draw[->] (A) -- (D);
						\draw[->] (B) -- (E);
						\draw[->] (B) -- (F);
						\draw[densely dotted, blue, line width=0.3mm, rounded corners=2pt] ($(A) + (-0.4,-0.4)$) rectangle ($(A) + (0.4,0.4)$);
						\draw[densely dotted, blue, line width=0.3mm, rounded corners=2pt] ($(B) + (-0.4,-0.4)$) rectangle ($(B) + (0.4,0.4)$);
						\draw[densely dotted, blue, line width=0.3mm, rounded corners=2pt] ($(C) + (-0.45,-0.45)$) rectangle ($(C) + (0.45,0.45)$);
						\draw[densely dotted, blue, line width=0.3mm, rounded corners=2pt] ($(D) + (-0.45,-0.45)$) rectangle ($(D) + (0.45,0.45)$);
						\draw[densely dotted, blue, line width=0.3mm, rounded corners=2pt] ($(E) + (-0.45,-0.45)$) rectangle ($(E) + (0.45,0.45)$);
						\draw[densely dotted, blue, line width=0.3mm, rounded corners=2pt] ($(F) + (-0.45,-0.45)$) rectangle ($(F) + (0.45,0.45)$);
				\end{tikzpicture}}
				\label{fig:lcn}
			\end{subfigure}%
			
			\hspace{-0.1em}
			\captionsetup[subfigure]{position=above}
			\begin{subfigure}[b]{0.22\textwidth}
				\centering
				\caption{\hspace{0.05cm} Local Classifier \\ per Parent  \\ Node (\textbf{LCPN})}
				\raisebox{-0.5\height}{
					\begin{tikzpicture}[node distance=0.2cm, scale=0.4]
						\node[scale=0.5, font=\Large] (R) at (0, 0) {R};
						\node[scale=0.5, font=\Large, below right=of R] (B) {B};
						\node[scale=0.5, font=\Large, below left=of R] (A) {A};
						\node[scale=0.5, font=\Large, below left=of A, yshift=-0.5cm, xshift=0.5cm] (C) {C};
						\node[scale=0.5, font=\Large, below right=of A, yshift=-0.5cm, xshift=-0.5cm] (D) {D};                
						\node[scale=0.5, font=\Large, below left=of B, yshift=-0.5cm, xshift=0.5cm] (E) {E};
						\node[scale=0.5, font=\Large, below right=of B, yshift=-0.5cm, xshift=-0.5cm] (F) {F};            
						\draw[->] (R) -- (A);
						\draw[->] (R) -- (B);
						\draw[->] (A) -- (C);
						\draw[->] (A) -- (D);
						\draw[->] (B) -- (E);
						\draw[->] (B) -- (F);
						\draw[densely dotted, blue, line width=0.3mm, rounded corners=2pt] ($(A) + (-0.45,-0.45)$) rectangle ($(A) + (0.45,0.45)$);
						\draw[densely dotted, blue, line width=0.3mm, rounded corners=2pt] ($(B) + (-0.45,-0.45)$) rectangle ($(B) + (0.45,0.45)$);
						\draw[densely dotted, blue, line width=0.3mm, rounded corners=2pt] ($(R) + (-0.45,-0.45)$) rectangle ($(R) + (0.45,0.45)$);
					\end{tikzpicture}
				} 
				\label{fig:lcpn}
			\end{subfigure}%
			\hspace{-0.3em} 
			
			\captionsetup[subfigure]{position=above}
			\begin{subfigure}[b]{0.22\textwidth}
				\centering
				\caption{Local Classifier \\ per Level \textbf{LCL}}
				\raisebox{-0.5\height}{%
					\begin{tikzpicture}[node distance=0.2cm, scale=0.4]
						\node[scale=0.5, font=\Large] (R) at (0, 0) {R};
						\node[scale=0.5, font=\Large, below right=of R] (B) {B};
						\node[scale=0.5, font=\Large, below left=of R] (A) {A};
						\node[scale=0.5, font=\Large, below left=of A, yshift=-0.5cm, xshift=0.5cm] (C) {C};
						\node[scale=0.5, font=\Large, below right=of A, yshift=-0.5cm, xshift=-0.5cm] (D) {D};                
						\node[scale=0.5, font=\Large, below left=of B, yshift=-0.5cm, xshift=0.5cm] (E) {E};
						\node[scale=0.5, font=\Large, below right=of B, yshift=-0.5cm, xshift=-0.5cm] (F) {F};                       
						\draw[->] (R) -- (A);
						\draw[->] (R) -- (B);
						\draw[->] (A) -- (C);
						\draw[->] (A) -- (D);
						\draw[->] (B) -- (E);
						\draw[->] (B) -- (F);
						\draw[densely dotted, blue, line width=0.3mm, rounded corners=2pt] ($(A) + (-0.5,0.5)$) rectangle ($(B) + (0.5,-0.5)$);
						\draw[densely dotted, blue, line width=0.3mm, rounded corners=2pt] ($(C) + (-0.5,-0.5)$) rectangle ($(F) + (0.5,0.5)$);
				\end{tikzpicture}}
				\label{fig:lcl}
			\end{subfigure}%
			\hspace{-0.4em}
			
			\captionsetup[subfigure]{position=above}
			\begin{subfigure}[b]{0.22\textwidth}
				\centering
				\caption{ \\ Global \\ Classifier \\ }
				\centering
				\raisebox{-0.5\height}{%
					\begin{tikzpicture}[node distance=0.2cm, scale=0.4]
						\node[scale=0.5, font=\Large] (R) at (0, 0) {R};
						\node[scale=0.5, font=\Large, below right=of R, xshift=1] (B) {B};
						\node[scale=0.5, font=\Large, below left=of R, xshift=-1] (A) {A};
						\node[scale=0.5, font=\Large, below left=of A, yshift=-0.5cm, xshift=0.5cm] (C) {C};
						\node[scale=0.5, font=\Large, below right=of A, yshift=-0.5cm, xshift=-0.5cm] (D) {D};               
						\node[scale=0.5, font=\Large, below left=of B, yshift=-0.5cm, xshift=0.5cm] (E) {E};
						\node[scale=0.5, font=\Large, below right=of B, yshift=-0.5cm, xshift=-0.1cm] (F) {F};               
						\draw[->] (R) -- (A);
						\draw[->] (R) -- (B);
						\draw[->] (A) -- (C);
						\draw[->] (A) -- (D);
						\draw[->] (B) -- (E);
						\draw[->] (B) -- (F);
						\draw[densely dotted, blue, line width=0.3mm, rounded corners=2pt] ($(A) + (-0.75,-2)$) rectangle ($(F) + (0.75,2)$);
				\end{tikzpicture}}
				\label{fig:global_classifier}
			\end{subfigure}
		}
		\caption{Most popular types of classifiers.}
		\label{fig:local_classifiers}
	\end{figure}
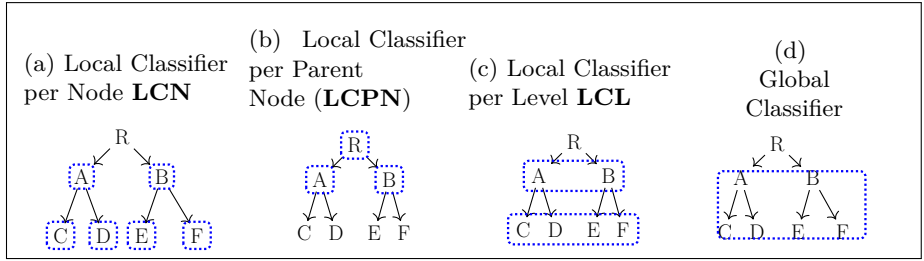

	Another important aspect to consider when developing a hierarchical classifier is the way in which the structure is navigated. The most frequently used local classifiers, shown in Figure \ref{fig:local_classifiers} are:
	
	\begin{itemize}[parsep=1pt]
		\setlength{\itemsep}{0.5pt}
		\setlength{\parskip}{0pt}
		\setlength{\parsep}{0pt}
		\item Local Classifier per Node (LCN): applies a binary classifier for each node of the hierarchy (Figure \ref{fig:lcn}), 
		\item Local Classifier per Parent Node (LCPN): assigns a separate multi-class classifier for each parent node (Figure \ref{fig:lcpn}), and 
		\item Local Classifier per Level (LCL): assigns one multi-class classifier for each hierarchical level (Figure \ref{fig:lcl}). 
	\end{itemize}

	To reduce model complexity and improve generalization LCL was chosen over the other alternatives. However, for the current level the predictions are not restricted solely to the subset of classes predicted at the previous level, thus giving the model the chance to learn and correct potential misclassifications at deeper levels. At each hierarchical level, the number of classes corresponds to the number of categories (nodes). With deeper advancement into the hierarchy, the categories become increasingly refined, leading to a larger number of classes. Ultimately, at the final level, all classes are included analogous to a flat classifier. 
	
\begin{figure}[htb!]
		\centering
		
		\begin{subfigure}{0.95\linewidth}
			\centering
			\fbox{\rule{0pt}{0.5in}
				\includegraphics[width=\linewidth]{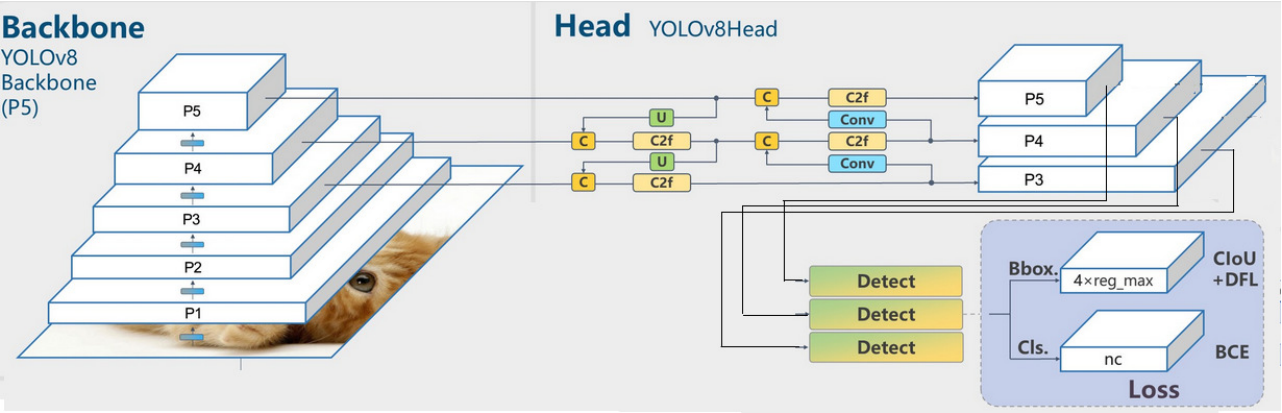}
			}
			\caption{YOLOv8 head architecture, illustrating the components responsible for bounding box regression and classification.}
			\label{fig:sub_orig_yolo_architecture}
		\end{subfigure}
		
		\vspace{1em} 
		
		\begin{subfigure}{0.95\linewidth}
			\centering
			\fbox{\rule{0pt}{0.5in}
				\includegraphics[width=\linewidth]{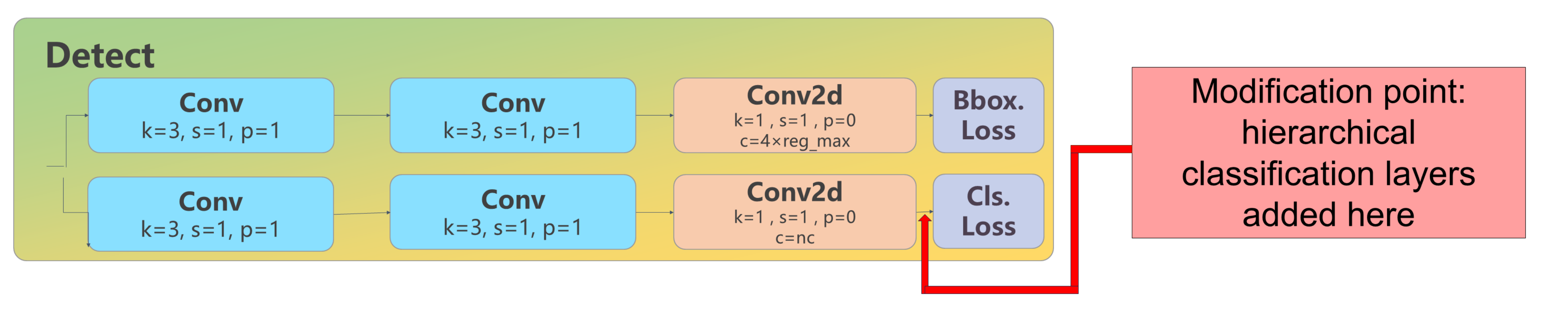}
			}
			\caption{Detection module of \YOLOvEight, annotated with a red arrow to highlight the designated insertion point for the proposed hierarchical classification layers.}
			\label{fig:sub_detect_module}
		\end{subfigure}
		
		\caption{Overview of YOLOv8 head and detection modules, adapted from \href{https://github.com/RangeKing/DL-Diagram/tree/cv}{DL-Diagram repository} illustrating both the architecture and proposed insertion points for hierarchical classification layers.}
	\end{figure}

	To support this hierarchical classification scheme, specific architectural modifications were made to the \YOLOvEight framework. Figure \ref{fig:sub_orig_yolo_architecture} illustrates the head module of the \YOLOvEight architecture (\citet{DBLP:journals/corr/abs-2408-15857}), which is responsible for producing the final predictions: the bounding box coordinates, confidence scores, and class labels. In addition, Figure \ref{fig:sub_detect_module} provides a detailed breakdown of the ``Detect'' component, explicitly highlighting the exact location within the classification branch where the hierarchical layers were integrated. The bounding box prediction pipeline remains unaltered; only the classification pathway is modified to accommodate the new hierarchical structure.

	To systematically investigate the optimal strategy for integrating hierarchical information, six alternative hierarchical architectures were explored, each differing in two main aspects: (1) the inserting point, where the hierarchical layers are inserted in the network, and (2) the concatenation point, where outputs from the preceding hierarchical level are merged with the current level. This variation was designed to identify the most effective point within the classification head at which hierarchical information should be integrated — ensuring that the signal passed from the preceding level is both maximally informative and optimally propagated backward during training, thus improving the flow of gradients and feature learning across hierarchical levels. The insertion point, shown in red in Figure \ref{fig:sub_orig_yolo_architecture} applies to versions $1$, $2$, $4$, and $6$. In versions $3$ and $5$, the hierarchical layers are inserted at earlier stages in the classification branch. A full comparison of the architectures is provided in the Experiments Section ~\ref{hierarchical_architectures_variants}. 
	
\begin{figure*}
		\centering
		\begin{subfigure}{0.8\textwidth}
			\centering
			\hspace*{-1.5cm}
			\resizebox{\textwidth}{!}{
				\begin{tikzpicture}[>=stealth,auto, node distance=2cm]
					\tikzstyle{block} = [rectangle, draw, minimum width=1.5cm, minimum height=1cm, align=center]
					\tikzstyle{virtual} = [coordinate]
					\tikzstyle{smallblock} = [rectangle, draw, minimum width=1cm, minimum height=1cm, align=center]
					\tikzstyle{parallelogramblock} = [draw, trapezium, very thin, trapezium left angle=85, trapezium right angle=95, minimum width=0.5cm, minimum height=1.3cm, text centered, text width=1.5cm, rounded corners, fill=mygray!15]
					\newcommand{\figthree}{ 
						\node [virtual] (input1) {};
						\node [block, right of=input1, node distance=2.5cm] (Conv1_1) {Conv \\ k=3, s=1, \\ p=1};
						\node [virtual, right of=Conv1_1] (output1_1) {};
						\node [block, right of=output1_1, node distance=0.5cm] (Conv2_1) {Conv \\ k=3, s=1, \\ p=1};
						\node [virtual, right of=Conv2_1] (output2_1) {};
						\node [parallelogramblock, right of=output2_1, node distance=0.5cm] (Conv3_1) {Conv2d \\ k=3, s=1, \\ p=1, c=nc};
						\node [virtual, right of=Conv3_1] (output3_1) {};
						\node [smallblock, right of=output3_1, node distance=0.5cm] (cls_loss_1) {cls \\ loss};
						\node [virtual] (input2) {};
						\node [virtual, below of=input1, node distance=2.5cm] (input2) {};
						\node [block, right of=input2, node distance=2.5cm] (Conv1_2) {Conv \\ k=3, s=1, \\ p=1};
						\node [virtual, right of=Conv1_2] (output1_2) {};
						\node [block, right of=output1_2, node distance=0.5cm] (Conv2_2) {Conv \\ k=3, s=1, \\ p=1};
						\node [virtual, right of=Conv2_2] (output2_2) {};
						\node [parallelogramblock, right of=output2_2, node distance=0.5cm] (Conv3_2) {Conv2d \\ k=3, s=1, \\ p=1, c=nc};
						\node [virtual, right of=Conv3_2] (output3_2) {};
						\node [parallelogramblock, right of=output3_2, node distance=0.8cm] (Conv3_3) {Conv2d \\ k=3, s=1, \\ p=1, c=nc};
						\node [virtual, right of=Conv3_3] (output3_3) {};
						\node [smallblock, right of=output3_3, node distance=0.3cm] (cls_loss_2) {cls \\ loss};
						\node [below=0.05cm of input1, align=center, xshift=0.7cm] {level 0};
						\node [below=0.05cm of input2, align=center, xshift=0.7cm] {level 1};
						\node [virtual] (input3) {};
						\node [virtual, below of=input2, node distance=2.5cm] (input3) {};
						\node [block, right of=input3, node distance=2.5cm] (Conv1_3) {Conv \\ k=3, s=1, \\ p=1};
						\node [virtual, right of=Conv1_3] (output1_3) {};
						\node [block, right of=output1_3, node distance=0.5cm] (Conv2_3) {Conv \\ k=3, s=1, \\ p=1};
						\node [virtual, right of=Conv2_3] (output2_3) {};
						\node [parallelogramblock, right of=output2_3, node distance=0.5cm] (Conv3_33) {Conv2d \\ k=3, s=1, \\ p=1, c=nc};
						\node [virtual, right of=Conv3_33] (output3_3) {};
						\node [parallelogramblock, right of=output3_3, node distance=0.8cm] (Conv3_4) {Conv2d \\ k=3, s=1, \\ p=1, c=nc};
						\node [virtual, right of=Conv3_4] (output3_4) {};
						\node [smallblock, right of=output3_4, node distance=0.3cm] (cls_loss_3) {cls \\ loss};
						\node [below=0.05cm of input1, align=center, xshift=0.7cm] {level 0};
						\node [below=0.05cm of input2, align=center, xshift=0.7cm] {level 1};
						\node [below=0.05cm of input3, align=center, xshift=0.7cm] {level 2};                   
						\draw[->] ([xshift=-1cm]input1) -- node[align=center] {input from\\backbone} (Conv1_1);
						\draw [->] ([xshift=-1cm]input2) -- node[align=center] {input from\\backbone} (Conv1_2);
						\draw [->] ([xshift=-1cm]input3) -- node[align=center] {input from\\backbone} (Conv1_3);
						\draw [->] (Conv1_1) -- (Conv2_1);
						\draw [->] (Conv2_1) -- (Conv3_1);
						\draw [->] (Conv3_1) -- (cls_loss_1);
						\draw [->] (Conv1_2) -- (Conv2_2);
						\draw [->] (Conv2_2) -- (Conv3_2);
						\draw [->] (Conv3_2) -- (Conv3_3);
						\draw [->] (Conv3_3) -- (cls_loss_2);
						\draw [->] (Conv1_3) -- (Conv2_3);
						\draw [->] (Conv2_3) -- (Conv3_33);
						\draw [->] (Conv3_33) -- (Conv3_4);
						\draw [->] (Conv3_4) -- (cls_loss_3);
						\draw [->] [red, densely dashed, line width=1.5pt] ([xshift=0.6cm]Conv3_1.east) -- ++(0,-2.5cm);
						\draw [red, densely dashed, line width=1.5pt] ([xshift=0.6cm]Conv3_3.east) -- ++(0,-1.3cm) coordinate (firstTip);
						\draw [red, densely dashed, line width=1.5pt] (firstTip) -- ++(-3cm, 0) coordinate (secondTip);
						\draw [->] [red, densely dashed, line width=1.5pt] (secondTip) -- ++(0,-1.2cm);
					}
					\figthree
				\end{tikzpicture}
			} 
		\end{subfigure}
		\caption{Hierarchical architecture $V4$ implemented in \YOLOvEight.}
		\label{fig:hierarchical_yolo_architecture_V4}
	\end{figure*}
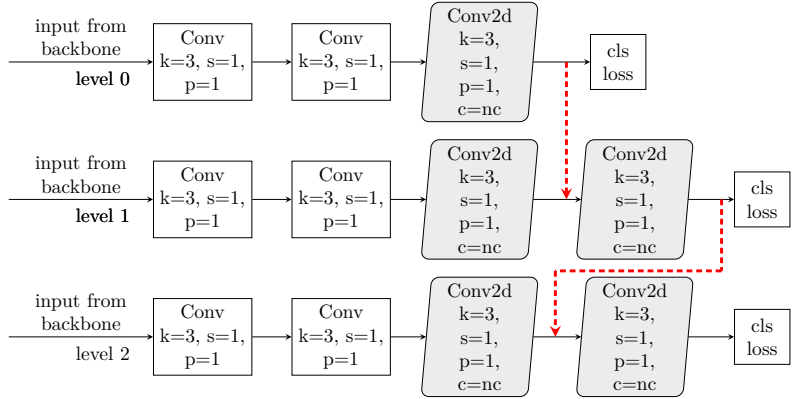

	Here, we focus our analysis on architecture \textit{Version 4} (Figure \ref{fig:hierarchical_yolo_architecture_V4}), which demonstrated the highest classification performance across evaluation metrics. To better understand its design and functionality, consider a hierarchical architecture consisting of three distinct levels, where the classification tasks are distributed as follows: \textit{level 0} predicts $2$ classes, \textit{level 1} predicts $10$ classes, and \textit{level 3} predicts $20$ classes.
	
	First, the initial input, derived from the backbone of the \YOLOvEight model, is replicated three times, thereby providing an independent input stream for each hierarchical level. 
	
	At \textit{level 0}, the Conv2D layer’s number of output channels corresponds directly to the $2$ target classes. Since only information specific to this current level is available at this stage, classification is performed in a flat manner, consistent with the standard approach employed in \YOLOvEight. For \textit{level 1}, the input is again sourced from the backbone; however, after the first Conv2D operation, feature information from the preceding hierarchical level (\textit{level 0}) is integrated. This fusion allows \textit{level 1} to incorporate the results from \textit{level 0} alongside its own features, thereby embedding contextual information from the preceding classification stage.
	
	Moreover, a second Conv2D layer is introduced at \textit{level 1} (and at each subsequent level) to refine the combined feature representation. This supplementary layer serves to adapt and transform the merged inputs: both the raw backbone features and the propagated information from the previous level, addressing the increased number of classes at this stage. By doing so, the network ensures that the hierarchical information is effectively used and that the subsequent predictions at \textit{level 1} are informed by prior classification results.
	
	This mechanism of information propagation continues recursively through the hierarchy. Each successive level receives not only the original input from the backbone but also enriched feature representations that encapsulate the predictions from all preceding levels. Since the bounding boxes remain constant across the hierarchical levels, the same physical object is progressively classified with increasing specificity: for example, it may be identified as ``food'' at \textit{level 0}, refined to ``bottle'' at \textit{level 1}, and further specified as ``bottle of milk'' at \textit{level 2}. This hierarchical classification framework thus enables the model to leverage shared spatial information while incrementally enriching the semantic detail of the predictions at each successive level.

	\subsection{Hierarchical Models Evaluation Metrics}\label{set_metric}
	Metrics such as precision, recall, accuracy, or $F1$ score are commonly utilized when evaluating a classification model (\cite{DBLP:journals/corr/abs-2209-05355}, \cite{Tharwat2020ClassificationAM}). However, these traditional metrics lack the ability to discriminate between different types of misclassification errors in hierarchical classification, since they ignore the relationships between classes (\cite{DBLP:conf/ai/KiritchenkoMNF06}, \cite{DBLP:journals/datamine/KosmopoulosPGPA15}, \cite{DBLP:journals/corr/abs-2306-09461}). In hierarchical classification scenarios, it is often more desirable to misclassify an instance into a proximate category rather than a distant one. To address this challenge and effectively assess the model performance of the architectures described in the previous section, we adopted the hierarchical metric proposed by \cite{DBLP:conf/ai/KiritchenkoMNF06}. A fundamental concept in this metric is the set of ancestors of a given node $C$, which comprises all nodes lying on paths leading to $C$, excluding the root node. This metric assesses the distance between the actual class and the prediction, taking into account the number of their common ancestors. It distinguishes between misclassifications in the same subgraph or in a remote subgraph and applies penalties accordingly.
	
	Thus, the conventional evaluation metrics, namely precision, recall, and $F_\beta$ score, are modified to capture the hierarchical structure of the classification. The modified hierarchical precision ($\text{Prec}_{Hier}$), hierarchical recall ($\text{Rec}_{Hier}$), and hierarchical $F_\beta$ ($F_{\beta, \text{Hier}}$) are defined with the following equations:
	
	\begin{equation}
		\text{Prec}_{Hier} = \frac {\mid \text{Ancest}(C_p) \cap \text{Ancest}(C_t)\mid}{\mid \text{Ancest}(C_p) \mid}
		\label{eq:hier_precision}
	\end{equation}
	
	\begin{equation}
		\text{Rec}_{Hier} = \frac {\mid \text{Ancest}(C_p) \cap \text{Ancest}(C_t)\mid}{\mid \text{Ancestor}(C_t) \mid}
		\label{eq:hier_recall}
	\end{equation}
	
	\begin{equation}
		F_{\beta, \text{Hier}} = \frac {\mid (\beta^2 +1) * \text{Prec}_{Hier} *  \text{Rec}_{Hier} \mid}{\mid (\beta^2 * \text{Prec}_{Hier} +  \text{Rec}_{Hier}) \mid}
		\label{eq:hier_F1}
	\end{equation}
	where:
	\begin{itemize}[parsep=1pt]
		\setlength{\itemsep}{0.5pt}
		\setlength{\parskip}{0pt}
		\setlength{\parsep}{0pt}
		
		\item[] $C_p$ is the predicted class
		\item[] $C_t$ is the ground-truth class
		\item[] $\beta \in [0, \infty)$, by default $\beta = 1$
		\item[] the $|\dots|$ denotes the number of elements in the set
	\end{itemize}
	
	$\text{Prec}_{Hier}$ and $\text{Rec}_{Hier}$ are calculated based on the number of common ancestors between the predicted and ground-truth classes. This value is then divided either by the total number of ancestors of the predicted class or by the total number of ancestors of the ground-truth class. The hierarchical $F_\beta$ score allows flexible weighting of $\text{Prec}_{Hier}$ and $\text{Rec}_{Hier}$. In the most commonly used case, when $\beta = 1$, the measure reduces to the hierarchical $F1$ score ($F1_{Hier}$), which assigns equal importance to precision and recall.
	
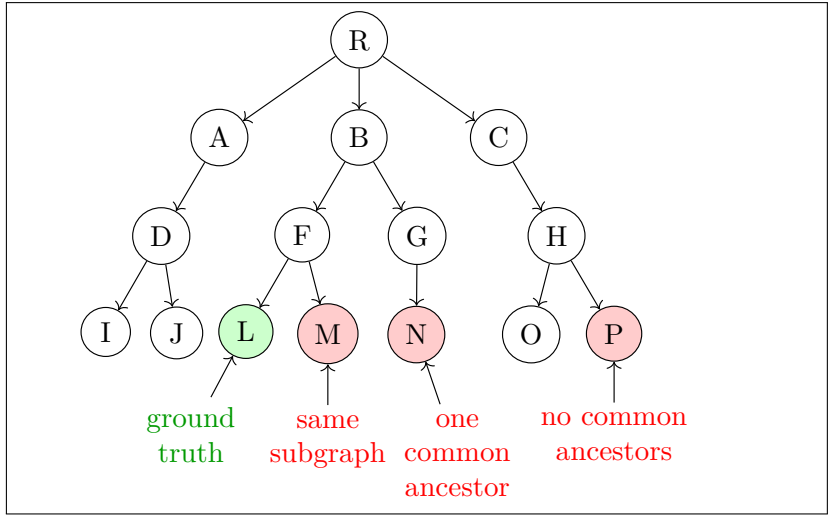
\begin{figure}[htb!]
		\centering
		\fbox{\rule{0pt}{0.8in}
			\begin{minipage}{0.8\columnwidth}
				\hspace*{0.5cm}
				\resizebox{0.8\linewidth}{!}{
					\begin{tikzpicture}[node distance=1cm, scale=0.8]
						\node[draw, circle] (R) at (0, 0) {R};
						\node[draw, circle, below=0.5cm of R] (B) {B};
						\node[draw, circle, left=1cm of B] (A) {A};
						\node[draw, circle, right=1cm of B] (C) {C};
						\node[draw, circle, below left=of A, xshift=0.5cm] (D) {D};
						\node[draw, circle, below left=of B, xshift=0.5cm] (F) {F};
						\node[draw, circle, below right=of B, xshift=-0.5cm] (G) {G};            
						\node[draw, circle, below right=of C, xshift=-0.5cm] (H) {H};
						\node[draw, circle, below left =of D, xshift=0.5cm] (I) {I};
						\node[draw, circle, below right=of D, xshift=-1cm] (J) {J};             
						\node[draw, circle, fill=green!20, below left=of F, xshift=0.5cm] (L) {L};
						\node[draw, circle, fill=red!20, below right=of F, xshift=-0.90cm] (M) {M};
						\node[draw, circle, fill=red!20, below left=1cm of G, xshift=1.2cm] (N) {N};
						\node[draw, circle, below left=of H, xshift=0.9cm] (O) {O};
						\node[draw, circle, fill=red!20, below right=of H, xshift=-0.5cm] (P) {P};
						\draw[->] (R) -- (A);
						\draw[->] (R) -- (B);
						\draw[->] (R) -- (C);
						\draw[->] (A) -- (D);
						\draw[->] (B) -- (F);
						\draw[->] (B) -- (G);
						\draw[->] (C) -- (H);
						\draw[->] (D) -- (I);
						\draw[->] (D) -- (J);
						\draw[->] (F) -- (L);
						\draw[->] (F) -- (M);
						\draw[->] (G) -- (N);
						\draw[->] (H) -- (O);
						\draw[->] (H) -- (P);
						\node[text=mygreen, align=center, below left=0.8cm of L, xshift=0.8cm] (groundtrue) {ground \\ truth};
						\draw[->] (groundtrue) -- (L);
						\node[text=red, align=center, below=0.5cm of M] (samesubgraph) {same \\ subgraph};
						\draw[->] (samesubgraph) -- (M);
						\node[text=red, align=center, below=0.5cm of N, xshift=0.5cm] (onecommonancestor) {one \\ common \\ ancestor};
						\draw[->] (onecommonancestor) -- (N);
						\node[text=red, align=center, below=0.5cm of P] (nocommonancestors) {no common \\ ancestors};
						\draw[->] (nocommonancestors) -- (P);
					\end{tikzpicture}
				}
		\end{minipage}}
		\caption{A tree structured hierarchy.}
		\label{fig:hierarchical_structure}
	\end{figure}

	As an illustration, examine the hierarchy in Figure \ref{fig:hierarchical_structure}, where the ground-truth node $C_t = L$. Consider three potential scenarios for misclassification: 
	
	\begin{enumerate}[parsep=1pt]
		\setlength{\itemsep}{0.5pt}
		\setlength{\parskip}{0pt}
		\setlength{\parsep}{0pt}
		
		\item If the predicted node $C_p = M$ is in the same subgraph as the ground-truth node, $\text{Ancest}(C_p) = \text{Ancest}(M) = \{B,F,M \}$, $\text{Ancest}(C_p) \cap \text{Ancest}(C_t) =\{B,F\}$, and the number of elements in the intersection $|\text{Ancest}(C_p) \cap \text{Ancest}(C_t)| =|\{B,F\}| = 2$. The total number of ancestors for $C_t$ is $|\{B,F,L\}|=3$, and for $C_p = M$ is $|\text{Ancest}(M)| = |\{B,F,M \}| = 3$. Then $\text{Prec}_M = \text{Rec}_M = F1_M = 2/3$. 
		
		\item If the predicted node $C_p = N$ is in a different subgraph with one common ancestor with the ground truth, $\text{Ancest}(C_p) = \text{Ancest}(N) = \{B,G,N\}$, $\text{Ancest}(C_p) \cap \text{Ancest}(C_t) =\{B\}$, and the number of elements in the intersection $|\text{Ancest}(C_p) \cap \text{Ancest}(C_t)| =|\{B\}| = 1$.) The total number of ancestors for $C_p = N$ is $|\text{Ancest}(N)| = |\{B,G,N \}| = 3$. Then $\text{Prec}_N = \text{Rec}_N = F1_N = 1/3$. 
		
		\item If the predicted node $C_p = P$ and the ground-truth node do not share any common ancestors, all $\text{Prec}_{Hier}$, $\text{Rec}_{Hier}$ and $F1_{Hier}$ for $P$ are $0$.
	\end{enumerate}
	
	If a flat classifier is used, all three scenarios result in metrics of $0$, regardless of how closely the prediction aligns with the ground truth in terms of hierarchy. In contrast, this modified metric imposes lesser penalties for misclassifications occurring within the same subgraph as the ground truth (mistaking $M$ for $L$) compared to misclassifications across different and more distant subgraphs (mistaking $L$ for $N$ or $P$).
	
	The hierarchical architectures outlined in the preceding subsection were evaluated using this novel metric. Among them, $V4$ (Figure \ref{fig:hierarchical_yolo_architectures_V4_V5_V6}\textcolor{red}{a}) showed the highest $F1_{Hier}$ score.

	\subsection{Modified Loss Function}\label{modified_loss}
	The loss function used in \YOLOvEight consists of two key components: a classification term (the standard Binary Cross Entropy Loss), and a bounding box (regression) term. The latter itself is a combination of two independent losses: Distribution Focal Loss (DFL) proposed by \cite{DBLP:conf/nips/0041WW00LT020} and Complete Intersection over Union (CIoU) loss presented by \cite{DBLP:conf/aaai/ZhengWLLYR20}. Originally designed to address the class imbalance problem in object detection tasks, in \YOLOvEight DFL is also used to improve bounding box regression, especially for difficult to predict objects with blurry or unclear boundaries. On the other hand, CIoU loss considers the aspect ratio differences between the predicted and ground-truth boxes in addition to the overlap between them. The final loss is a weighted sum of these three individual losses (Equation \ref{eq:total_loss}).
	
	\begin{subequations}
		\begin{align}
			&\mathcal{L} = \mathcal{L}_{\text{Reg}} + \mathcal{L}_{\text{Cls}} \label{eq:total_loss} \\
			&\mathcal{L}_{\text{Reg}} = w_{\text{box}} \cdot \mathcal{L}_{\text{CIoU}} + w_{\text{dfl}} \cdot \mathcal{L}_{\text{DFL}} \\
			&\mathcal{L}_{\text{Cls}} = w_{\text{cls}} \cdot \mathcal{L}_{\text{BCE}}
		\end{align}
	\end{subequations}
	where: 
	\begin{itemize}
		\setlength{\itemsep}{0.5pt}
		\setlength{\parskip}{0pt}
		\setlength{\parsep}{0pt}
		
		\item[] $w_{\text{box}}$ -- weight of CIoU loss with default value of $7.5$ in \YOLOvEight
		\item[] $w_{\text{dfl}}$ -- \xspace weight of DFL loss with default value of $1.5$ in \YOLOvEight
		\item[] $w_{\text{cls}}$ -- \xspace weight of BCE loss with default value of $0.5$ in \YOLOvEight
	\end{itemize}
	
	To train the neural network to follow the hierarchical structure during object classification, modifications were made to \YOLO's loss function. First, for each hierarchical level a distinct loss function is calculated. It is the weighted sum of the individual CIoU, DFL and BCE losses for this particular level. The weights $w_{\text{box}}$, $w_{\text{dfl}}$, and $w_{\text{cls}}$ are constants and set to their default values at each level to maintain a consistent contribution from each loss component to the total loss. Finally, the total loss for the entire hierarchical structure is defined as the average of these level-specific losses. This ensures that the model is trained comprehensively across all hierarchical levels.
	
	\begin{equation}
		\begin{aligned}
			\mathcal{L} = \sum_{i=0}^{l} (\mathcal{L}_{Regression_i} + \mathcal{L}_{Class_i}) \\
			\label{eq:yolov_hier_loss}
		\end{aligned}
	\end{equation}
	where: 
	\begin{itemize}[parsep=1pt]
		\item[] $l$ -- hierarchy depth, number of hierarchical levels.
	\end{itemize}
	
	Furthermore, an additional term to penalize predictions which are not children of the parent at the previous hierarchical level was incorporated in the classification loss. As a result, the modified loss function is hierarchy-aware, penalizing both classification errors and violations of the hierarchy.
	
	For the first hierarchical level no penalty term is applied, since there are no parent classes for this level. 
	
	\begin{equation}
		\mathcal{L}_{Cls_0} = w_{\text{cls}} \cdot \mathcal{L}_{BCE_0}
		\label{eq:modified_loss1}
	\end{equation}
	
	For each subsequent hierarchical level if the predicted class is a child of the parent class on the previous hierarchical level, no penalty is applied. However, if the predicted class is not a child of the parent node, the penalty term is set to be the confidence score of the predicted class at the current level. Consequently, the modified classification loss is computed as follows:
	
	%

\begin{equation}
	\mathcal{L}_{Cls_l} = w_{\mathrm{cls}} \cdot \left( \mathcal{L}_{BCE_l} + \alpha \cdot \sum_{i=1}^{S} (1 - \delta_{il}) \cdot \mathrm{conf}_{il} \right)
	\label{eq:modified_loss2}
\end{equation}
where: 
\begin{itemize}[parsep=1pt]
	\setlength{\itemsep}{0.5pt}
	\setlength{\parskip}{0pt}
	\setlength{\parsep}{0pt}
	
	\item[] $l$ -- hierarchy depth, number of hierarchical levels
	\item[] $S$ -- number of classes
	\item[] $\mathcal{L}_{BCE_l}$ -- Binary Cross Entropy at hierarchical \textit{level l}
	\item[] $\alpha$ -- regularization constant $\alpha \geq 0$
	\item[]
	$
	\delta_{il} =
	\begin{cases}
		1, & \text{if class } i \text{ is a child of its parent at level } l-1 \\
		0, & \text{otherwise}
	\end{cases}
	$
\end{itemize}

Penalizing an incorrect prediction with its confidence score has an additional advantage: the higher the confidence of the incorrect prediction, the harsher the penalty. Conversely, for incorrect predictions with lower confidence the penalty is less severe. If the regularization constant is set to $\alpha = 0$, the penalty term is completely omitted. In this case, the loss function reduces to the standard BCE loss, and no hierarchical consistency enforcement is applied at that level.

\section{Experiments}\label{experiments}

\subsection{Experimental Variants of Hierarchical Architectures} \label{hierarchical_architectures_variants}

\begin{figure}[H]
	\centering
	\begin{subfigure}{0.8\textwidth} 
		\centering
		\resizebox{\textwidth}{!}{
			\begin{tikzpicture}[>=stealth,auto, node distance=1.5cm]
				\tikzstyle{block} = [rectangle, draw, minimum width=1.5cm, minimum height=1cm, align=center]
				\tikzstyle{virtual} = [coordinate]
				\tikzstyle{smallblock} = [rectangle, draw, minimum width=1cm, minimum height=1cm, align=center]
				\tikzstyle{parallelogramblock} = [draw, trapezium, very thin, trapezium left angle=85, trapezium right angle=95, minimum width=0.5cm, minimum height=1.3cm, text centered, text width=1.5cm, rounded corners, fill=mygray!15]
				
				\newcommand{\figone}{ 
					\node [virtual] (input) {};
					\node [block, right of=input, node distance=2.5cm, rounded corners] (Conv1)     {Conv \\ k=3, s=1, \\ p=1};
					\node [virtual, right of=Conv1] (output1) {};
					\node [block, right of=output1, node distance=0.5cm, rounded corners] (Conv2)    {Conv \\ k=3, s=1, \\ p=1};
					\node [virtual, right of=Conv2] (output2) {};
					\node [parallelogramblock, right of=output2, node distance=0.5cm] (Conv3)    {Conv2d \\ k=3, s=1, \\ p=1, c=nc};
					\node [virtual, right of=Conv3] (output3) {};
					\node [smallblock, right of=output3, node distance=0.2cm, rounded corners] (cls_loss) {cls \\ loss};
					\node [above=0.5cm of Conv2, align=center, font=\bfseries\large] (annotation) {Original \YOLOvEight Architecture};
					\draw [->] (input) -- node[align=center] {input from\\backbone} (Conv1);
					\draw [->] (Conv1) -- (Conv2);
					\draw [->] (Conv2) -- (Conv3);
					\draw [->] (Conv3) -- (cls_loss);
				}
				\figone
			\end{tikzpicture}
		} 
	\end{subfigure}
	\begin{subfigure}{\textwidth}
		\centering
		\resizebox{!}{0.78\textheight}{
			\begin{tikzpicture}[>=stealth,auto, node distance=1.5cm]
				\tikzstyle{block} = [rectangle, draw, minimum width=1.5cm, minimum height=1cm, align=center]
				\tikzstyle{virtual} = [coordinate]
				\tikzstyle{smallblock} = [rectangle, draw, minimum width=1cm, minimum height=1cm, align=center]
				\tikzstyle{parallelogramblock} = [draw, trapezium, very thin, trapezium left angle=85, trapezium right angle=95, minimum width=0.1cm, minimum height=1.3cm, text centered, text width=1.5cm, rounded corners, fill=mygray!15]
				
				\newcommand{\figtwo}{ 
					\node [virtual] (input1) {};
					\node [block, right of=input1, node distance=2.5cm] (Conv1_1) {Conv \\ k=3, s=1, \\ p=1};
					\node [virtual, right of=Conv1_1] (output1_1) {};
					\node [block, right of=output1_1, node distance=0.5cm] (Conv2_1) {Conv \\ k=3, s=1, \\ p=1};
					\node [virtual, right of=Conv2_1] (output2_1) {};
					\node [parallelogramblock, right of=output2_1, node distance=0.7cm] (Conv3_1) {Conv2d \\ k=3, s=1, \\ p=1, c=nc};
					\node [virtual, right of=Conv3_1] (output3_1) {};
					\node [smallblock, right of=output3_1, node distance=0.3cm] (cls_loss_1) {cls \\ loss};
					\node [virtual] (input2) {};
					\node [virtual, below of=input1, node distance=2.5cm] (input2) {};
					\node [block, right of=input2, node distance=2.5cm] (Conv1_2) {Conv \\ k=3, s=1, \\ p=1};
					\node [virtual, right of=Conv1_2] (output1_2) {};
					\node [block, right of=output1_2, node distance=0.5cm] (Conv2_2) {Conv \\ k=3, s=1, \\ p=1};
					\node [virtual, right of=Conv2_2] (output2_2) {};
					\node [parallelogramblock, right of=output2_2, node distance=0.7cm] (Conv3_2) {Conv2d \\ k=3, s=1, \\ p=1, c=nc};
					\node [virtual, right of=Conv3_2] (output3_2) {};
					\node [smallblock, right of=output3_2, node distance=0.3cm] (cls_loss_2) {cls \\ loss};
					\node [above=0.5cm of Conv2, align=center, font=\bfseries\large] (annotation) {Hierarchy Architecture V1};
					\node [below=0.05cm of input1, align=center, xshift=0.7cm] {level 0};
					\node [below=0.05cm of input2, align=center, xshift=0.7cm] {level 1};
					\draw [->] (input1) -- node[align=center] {input from\\backbone} (Conv1_1);
					\draw [->] (input2) -- node[align=center] {input from\\backbone} (Conv1_2);
					\draw [->] (Conv1_1) -- (Conv2_1);
					\draw [->] (Conv2_1) -- (Conv3_1);
					\draw [->] (Conv3_1) -- (cls_loss_1);
					\draw [->] (Conv1_2) -- (Conv2_2);
					\draw [->] (Conv2_2) -- (Conv3_2);
					\draw [->] (Conv3_2) -- (cls_loss_2);
					\draw[red, dashed, line width=1.5pt] ([xshift=0.3cm]Conv3_1.east) -- ++(0,-3.35cm);
					\draw[red, dashed, line width=1.5pt] (2.5cm,-3.3cm) -- ++(17cm,0);
					\draw [->] [red, dashed, line width=1.5pt] (2.5, -3.3) -- ++(0, -2.8);
				}
				\newcommand{\figfour}{ 
					\node [virtual] (input1) {};
					\node [block, right of=input1, node distance=2.5cm] (Conv1_1) {Conv \\ k=3, s=1, \\ p=1};
					\node [virtual, right of=Conv1_1] (output1_1) {};
					\node [block, right of=output1_1, node distance=0.5cm] (Conv2_1) {Conv \\ k=3, s=1, \\ p=1};
					\node [virtual, right of=Conv2_1] (output2_1) {};
					\node [parallelogramblock, right of=output2_1, node distance=0.7cm] (Conv3_1) {Conv2d \\ k=3, s=1, \\ p=1, c=nc};
					\node [virtual, right of=Conv3_1] (output3_1) {};
					\node [smallblock, right of=output3_1, node distance=0.3cm] (cls_loss_1) {cls \\ loss};
					\node [virtual] (input2) {};
					\node [virtual, below of=input1, node distance=2.5cm] (input2) {};
					\node [block, right of=input2, node distance=2.5cm] (Conv1_2) {Conv \\ k=3, s=1, \\ p=1};
					\node [virtual, right of=Conv1_2] (output1_2) {};
					\node [block, right of=output1_2, node distance=0.5cm] (Conv2_2) {Conv \\ k=3, s=1, \\ p=1};
					\node [virtual, right of=Conv2_2] (output2_2) {};
					\node [parallelogramblock, right of=output2_2, node distance=0.7cm] (Conv3_2) {Conv2d \\ k=3, s=1, \\ p=1, c=nc};
					\node [virtual, right of=Conv3_2] (output3_2) {};
					\node [smallblock, right of=output3_2, node distance=0.3cm] (cls_loss_2) {cls \\ loss};
					\node [above=0.5cm of Conv2, align=center, font=\bfseries\large] (annotation) {Hierarchy Architecture V2};
					\node [below=0.05cm of input1, align=center, xshift=0.7cm] {level 0};
					\node [below=0.05cm of input2, align=center, xshift=0.7cm] {level 1};
					\draw [->] (input1) -- node[align=center] {input from\\backbone} (Conv1_1);
					\draw [->] (input2) -- node[align=center] {input from\\backbone} (Conv1_2);
					\draw [->] (Conv1_1) -- (Conv2_1);
					\draw [->] (Conv2_1) -- (Conv3_1);
					\draw [->] (Conv3_1) -- (cls_loss_1);
					\draw [->] (Conv1_2) -- (Conv2_2);
					\draw [->] (Conv2_2) -- (Conv3_2);
					\draw [->] (Conv3_2) -- (cls_loss_2);
					\draw[red, densely dotted, line width=1.5pt] ([xshift=0.3cm]Conv3_1.east) -- ++(0,-3.35cm);
					\draw[red, densely dotted, line width=1.5pt] (13.7cm,-3.3cm) -- ++(5.8cm,0);
					\draw [->] [red, densely dotted, line width=1.5pt] (13.7, -3.3) -- ++(0, -2.8);
				}
				\newcommand{\figsix}{ 
					\node [virtual] (input1) {};
					\node [block, right of=input1, node distance=2.5cm] (Conv1_1) {Conv \\ k=3, s=1, \\ p=1};
					\node [virtual, right of=Conv1_1] (output1_1) {};
					\node [block, right of=output1_1, node distance=0.5cm] (Conv2_1) {Conv \\ k=3, s=1, \\ p=1};
					\node [virtual, right of=Conv2_1] (output2_1) {};
					\node [parallelogramblock, right of=output2_1, node distance=0.8cm] (Conv3_1) {Conv2d \\ k=3, s=1, \\ p=1, c=nc};
					\node [virtual, right of=Conv3_1] (output3_1) {};
					\node [smallblock, right of=output3_1, node distance=0.3cm] (cls_loss_1) {cls \\ loss};
					\node [virtual] (input2) {};
					\node [virtual, below of=input1, node distance=2.5cm] (input2) {};
					\node [block, right of=input2, node distance=2.5cm] (Conv1_2) {Conv \\ k=3, s=1, \\ p=1};
					\node [virtual, right of=Conv1_2] (output1_2) {};
					\node [block, right of=output1_2, node distance=0.5cm] (Conv2_2) {Conv \\ k=3, s=1, \\ p=1};
					\node [virtual, right of=Conv2_2] (output2_2) {};
					\node [parallelogramblock, right of=output2_2, node distance=0.8cm] (Conv3_2) {Conv2d \\ k=3, s=1, \\ p=1, c=nc};
					\node [virtual, right of=Conv3_2] (output3_2) {};
					\node [smallblock, right of=output3_2, node distance=0.3cm] (cls_loss_2) {cls \\ loss};
					\node [above=0.5cm of Conv2, align=center, font=\bfseries\large] (annotation) {Hierarchy Architecture V3};
					\node [below=0.05cm of input1, align=center, xshift=0.7cm] {level 0};
					\node [below=0.05cm of input2, align=center, xshift=0.7cm] {level 1};
					\draw [->] (input1) -- node[align=center] {input from\\backbone} (Conv1_1);
					\draw [->] (input2) -- node[align=center] {input from\\backbone} (Conv1_2);
					\draw [->] (Conv1_1) -- (Conv2_1);
					\draw [->] (Conv2_1) -- (Conv3_1);
					\draw [->] (Conv3_1) -- (cls_loss_1);
					\draw [->] (Conv1_2) -- (Conv2_2);
					\draw [->] (Conv2_2) -- (Conv3_2);
					\draw [->] (Conv3_2) -- (cls_loss_2);
					\draw [->] [red, densely dotted, line width=1.5pt] ([xshift=0.4cm]Conv1_1.east) -- ++(0,-6.3cm);
				}
				\node[scale =2,draw, thick, inner sep=5pt, rounded corners] (box) {%
					\begin{tikzpicture}
						\matrix[column sep=0.2cm, row sep=0.1cm] {
							\node (V1) {\begin{tikzpicture}[scale=0.4] \figtwo 
							\end{tikzpicture} }; \\
							\node (V2) {\begin{tikzpicture}[scale=0.4] \figfour \end{tikzpicture}};  \\
							\node (V3) {\begin{tikzpicture}[scale=0.4] \figsix \end{tikzpicture}}; \\
						};
						\node [right=-1.5cm of V1, yshift=2.25cm] {\textbf{(a)}\label{fig:V1}};
						\node [right=-1.5cm of V2, yshift=2.25cm] {\textbf{(b)}};
						\node [right=-1.5cm of V3, yshift=2.25cm] {\textbf{(c)}};
						\label{fig:v1_v3}
					\end{tikzpicture}
				};
				
			\end{tikzpicture}
		}
	\end{subfigure}
	\caption{Hierarchical architectures $V1$, $V2$ and $V3$ implemented in \YOLOvEight.}
	\label{fig:hierarchical_yolo_architectures_V1_V2_V3} 
\end{figure}
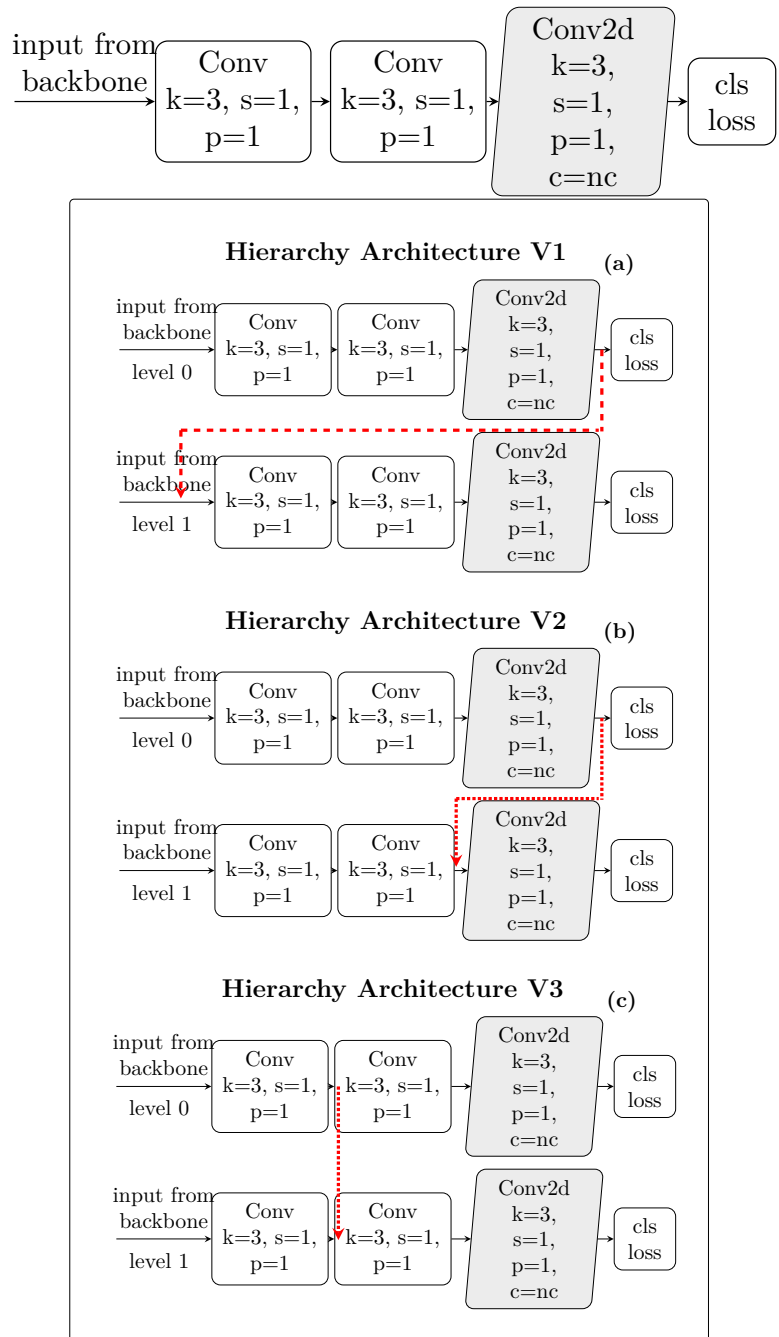

\begin{figure*}[!htbp]
	\centering
	\begin{subfigure}{0.8\textwidth}
		\centering
		\resizebox{\textwidth}{!}{
			\begin{tikzpicture}[>=stealth,auto, node distance=1.5cm]
				\tikzstyle{block} = [rectangle, draw, minimum width=1.5cm, minimum height=1cm, align=center]
				\tikzstyle{virtual} = [coordinate]
				\tikzstyle{smallblock} = [rectangle, draw, minimum width=1cm, minimum height=1cm, align=center]
				\tikzstyle{parallelogramblock} = [draw, trapezium, very thin, trapezium left angle=85, trapezium right angle=95, minimum width=0.5cm, minimum height=1.3cm, text centered, text width=1.5cm, rounded corners, fill=mygray!15]
				\newcommand{\figone}{ 
					\node [virtual] (input) {};
					\node [block, right of=input, node distance=2.5cm, rounded corners] (Conv1)     {Conv \\ k=3, s=1, \\ p=1};
					\node [virtual, right of=Conv1] (output1) {};
					\node [block, right of=output1, node distance=0.5cm, rounded corners] (Conv2)    {Conv \\ k=3, s=1, \\ p=1};
					\node [virtual, right of=Conv2] (output2) {};
					\node [parallelogramblock, right of=output2, node distance=0.5cm] (Conv3)    {Conv2d \\ k=3, s=1, \\ p=1, c=nc};
					\node [virtual, right of=Conv3] (output3) {};
					\node [smallblock, right of=output3, node distance=0.2cm, rounded corners] (cls_loss) {cls \\ loss};
					\node [above=0.5cm of Conv2, align=center, font=\bfseries\large] (annotation) {Original \YOLOvEight Architecture};
					\draw [->] (input) -- node[align=center] {input from\\backbone} (Conv1);
					\draw [->] (Conv1) -- (Conv2);
					\draw [->] (Conv2) -- (Conv3);
					\draw [->] (Conv3) -- (cls_loss);
				}
				\figone
			\end{tikzpicture}
		} 
	\end{subfigure}
	\begin{subfigure}{\textwidth} 
		\centering
		\resizebox{!}{0.78\textheight}{
			\begin{tikzpicture}[>=stealth,auto, node distance=1.5cm]
				\tikzstyle{block} = [rectangle, draw, minimum width=1.5cm, minimum height=1cm, align=center]
				\tikzstyle{virtual} = [coordinate]
				\tikzstyle{smallblock} = [rectangle, draw, minimum width=1cm, minimum height=1cm, align=center]
				\tikzstyle{parallelogramblock} = [draw, trapezium, very thin, trapezium left angle=85, trapezium right angle=95, minimum width=0.1cm, minimum height=1.3cm, text centered, text width=1.5cm, rounded corners, fill=mygray!15]
				\newcommand{\figthree}{ 
					\node [virtual] (input1) {};
					\node [block, right of=input1, node distance=2.5cm] (Conv1_1) {Conv \\ k=3, s=1, \\ p=1};
					\node [virtual, right of=Conv1_1] (output1_1) {};
					\node [block, right of=output1_1, node distance=0.5cm] (Conv2_1) {Conv \\ k=3, s=1, \\ p=1};
					\node [virtual, right of=Conv2_1] (output2_1) {};
					\node [parallelogramblock, right of=output2_1, node distance=0.5cm] (Conv3_1) {Conv2d \\ k=3, s=1, \\ p=1, c=nc};
					\node [virtual, right of=Conv3_1] (output3_1) {};
					\node [smallblock, right of=output3_1, node distance=0.5cm] (cls_loss_1) {cls \\ loss};
					\node [virtual] (input2) {};
					\node [virtual, below of=input1, node distance=2.5cm] (input2) {};
					\node [block, right of=input2, node distance=2.5cm] (Conv1_2) {Conv \\ k=3, s=1, \\ p=1};
					\node [virtual, right of=Conv1_2] (output1_2) {};
					\node [block, right of=output1_2, node distance=0.5cm] (Conv2_2) {Conv \\ k=3, s=1, \\ p=1};
					\node [virtual, right of=Conv2_2] (output2_2) {};
					\node [parallelogramblock, right of=output2_2, node distance=0.5cm] (Conv3_2) {Conv2d \\ k=3, s=1, \\ p=1, c=nc};
					\node [virtual, right of=Conv3_2] (output3_2) {};
					\node [parallelogramblock, right of=output3_2, node distance=0.8cm] (Conv3_3) {Conv2d \\ k=3, s=1, \\ p=1, c=nc};
					\node [virtual, right of=Conv3_3] (output3_3) {};
					\node [smallblock, right of=output3_3, node distance=0.3cm] (cls_loss_2) {cls \\ loss};
					\node [above=0.5cm of Conv2, align=center, font=\bfseries\large] (annotation) {Hierarchy Architecture V4};
					\node [below=0.05cm of input1, align=center, xshift=0.7cm] {level 0};
					\node [below=0.05cm of input2, align=center, xshift=0.7cm] {level 1};
					\draw [->] (input1) -- node[align=center] {input from\\backbone} (Conv1_1);
					\draw [->] (input2) -- node[align=center] {input from\\backbone} (Conv1_2);
					\draw [->] (Conv1_1) -- (Conv2_1);
					\draw [->] (Conv2_1) -- (Conv3_1);
					\draw [->] (Conv3_1) -- (cls_loss_1);
					\draw [->] (Conv1_2) -- (Conv2_2);
					\draw [->] (Conv2_2) -- (Conv3_2);
					\draw [->] (Conv3_2) -- (Conv3_3);
					\draw [->] (Conv3_3) -- (cls_loss_2);
					\draw [->] [red, densely dashed, line width=1.5pt] ([xshift=0.6cm]Conv3_1.east) -- ++(0,-6.3cm);
				}
				\newcommand{\figfive}{ 
					\node [virtual] (input1) {};
					\node [block, right of=input1, node distance=2.5cm] (Conv1_1) {Conv \\ k=3, s=1, \\ p=1};
					\node [virtual, right of=Conv1_1] (output1_1) {};
					\node [block, right of=output1_1, node distance=0.5cm] (Conv2_1) {Conv \\ k=3, s=1, \\ p=1};
					\node [virtual, right of=Conv2_1] (output2_1) {};
					\node [parallelogramblock, right of=output2_1, node distance=0.8cm] (Conv3_1) {Conv2d \\ k=3, s=1, \\ p=1, c=nc};
					\node [virtual, right of=Conv3_1] (output3_1) {};
					\node [smallblock, right of=output3_1, node distance=0.3cm] (cls_loss_1) {cls \\ loss};
					\node [virtual] (input2) {};
					\node [virtual, below of=input1, node distance=2.5cm] (input2) {};
					\node [block, right of=input2, node distance=2.5cm] (Conv1_2) {Conv \\ k=3, s=1, \\ p=1};
					\node [virtual, right of=Conv1_2] (output1_2) {};
					\node [block, right of=output1_2, node distance=0.5cm] (Conv2_2) {Conv \\ k=3, s=1, \\ p=1};
					\node [virtual, right of=Conv2_2] (output2_2) {};
					\node [parallelogramblock, right of=output2_2, node distance=0.8cm] (Conv3_2) {Conv2d \\ k=3, s=1, \\ p=1, c=nc};
					\node [virtual, right of=Conv3_2] (output3_2) {};
					\node [smallblock, right of=output3_2, node distance=0.3cm] (cls_loss_2) {cls \\ loss};
					\node [above=0.5cm of Conv2, align=center, font=\bfseries\large] (annotation) {Hierarchy Architecture V5};
					\node [below=0.05cm of input1, align=center, xshift=0.7cm] {level 0};
					\node [below=0.05cm of input2, align=center, xshift=0.7cm] {level 1};
					\draw [->] (input1) -- node[align=center] {input from\\backbone} (Conv1_1);
					\draw [->] (input2) -- node[align=center] {input from\\backbone} (Conv1_2);
					\draw [->] (Conv1_1) -- (Conv2_1);
					\draw [->] (Conv2_1) -- (Conv3_1);
					\draw [->] (Conv3_1) -- (cls_loss_1);
					\draw [->] (Conv1_2) -- (Conv2_2);
					\draw [->] (Conv2_2) -- (Conv3_2);
					\draw [->] (Conv3_2) -- (cls_loss_2);
					\draw [->] [red, densely dashed, line width=1.5pt] ([xshift=0.6cm]Conv2_1.east) -- ++(0,-6.3cm);
				}
				\newcommand{\figseven}{ 
					\node [virtual] (input1) {};
					\node [block, right of=input1, node distance=2.5cm] (Conv1_1) {Conv \\ k=3, s=1, \\ p=1};
					\node [virtual, right of=Conv1_1] (output1_1) {};
					\node [block, right of=output1_1, node distance=0.5cm] (Conv2_1) {Conv \\ k=3, s=1, \\ p=1};
					\node [virtual, right of=Conv2_1] (output2_1) {};
					\node [parallelogramblock, right of=output2_1, node distance=0.8cm] (Conv3_1) {Conv2d \\ k=3, s=1, \\ p=1, c=nc};
					\node [virtual, right of=Conv3_1] (output3_1) {};
					\node [smallblock, right of=output3_1, node distance=0.5cm] (cls_loss_1) {cls \\ loss};
					\node [virtual] (input2) {};
					\node [virtual, below of=input1, node distance=2.5cm] (input2) {};
					\node [block, right of=input2, node distance=2.5cm] (Conv1_2) {Conv \\ k=3, s=1, \\ p=1};
					\node [virtual, right of=Conv1_2] (output1_2) {};
					\node [block, right of=output1_2, node distance=0.5cm] (Conv2_2) {Conv \\ k=3, s=1, \\ p=1};
					\node [virtual, right of=Conv2_2] (output2_2) {};
					\node [parallelogramblock, right of=output2_2, node distance=0.8cm] (Conv3_2) {Conv2d \\ k=3, s=1, \\ p=1, c=nc};
					\node [virtual, right of=Conv3_2] (output3_2) {};
					\node [parallelogramblock, right of=output3_2, node distance=0.8cm] (Conv3_3) {Conv2d \\ k=3, s=1, \\ p=1, c=nc};
					\node [virtual, right of=Conv3_3] (output3_3) {};
					\node [smallblock, right of=output3_3, node distance=0.5cm] (cls_loss_2) {cls \\ loss};
					\node [virtual] (input3) {};
					\node [virtual, below of=input2, node distance=2.5cm] (input3) {};
					\node [block, right of=input3, node distance=2.5cm] (Conv1_3) {Conv \\ k=3, s=1, \\ p=1};
					\node [virtual, right of=Conv1_3] (output1_3) {};
					\node [block, right of=output1_3, node distance=0.5cm] (Conv2_3) {Conv \\ k=3, s=1, \\ p=1};
					\node [virtual, right of=Conv2_3] (output2_3) {};
					\node [parallelogramblock, right of=output2_3, node distance=0.8cm] (Conv3_3_3) {Conv2d \\ k=3, s=1, \\ p=1, c=nc};
					\node [virtual, right of=Conv3_3_3] (output3_3) {};
					\node [parallelogramblock, right of=output3_3, node distance=1cm] (Conv3_4) {Conv2d \\ k=3, s=1, \\ p=1, c=nc};
					\node [virtual, right of=Conv3_4] (output3_4) {};
					\node [smallblock, right of=output3_4, node distance=0.3cm] (cls_loss_3) {cls \\ loss};
					\node [above=0.5cm of Conv2, align=center, font=\bfseries\large] (annotation) {Hierarchy Architecture V6};
					\node [below=0.05cm of input1, align=center, xshift=0.7cm] {level 0};
					\node [below=0.05cm of input2, align=center, xshift=0.7cm] {level 1};
					\node [below=0.05cm of input3, align=center, xshift=0.7cm] {level 2};
					\draw [->] (input1) -- node[align=center] {input from\\backbone} (Conv1_1);
					\draw [->] (input2) -- node[align=center] {input from\\backbone} (Conv1_2);
					\draw [->] (input3) -- node[align=center] {input from\\backbone} (Conv1_3);
					\draw [->] (Conv1_1) -- (Conv2_1);
					\draw [->] (Conv2_1) -- (Conv3_1);
					\draw [->] (Conv3_1) -- (cls_loss_1);
					\draw [->] (Conv1_2) -- (Conv2_2);
					\draw [->] (Conv2_2) -- (Conv3_2);
					\draw [->] (Conv3_2) -- (Conv3_3);
					\draw [->] (Conv3_3) -- (cls_loss_2);
					\draw [->] (Conv1_3) -- (Conv2_3);
					\draw [->] (Conv2_3) -- (Conv3_3_3);
					\draw [->] (Conv3_3_3) -- (Conv3_4);
					\draw [->] (Conv3_4) -- (cls_loss_3);
					\draw [->] [red, densely dashed, line width=1.5pt] ([xshift=0.4cm]Conv3_1.east) -- ++(0,-6.3cm);
					\draw [red, densely dashed, line width=1.5pt] ([xshift=0.4cm]Conv3_3.east) -- ++(0,-3.35cm);
					\draw[red, densely dashed, line width=1.5pt] (19.9cm,-9.6cm) -- ++(5.8cm,0);
					\draw [red, densely dashed, line width=1.5pt] ([xshift=0.9cm]Conv3_1.east) -- ++(0,-3.35cm);
					\draw[red, densely dashed, line width=1.5pt] (20.3,-3.35cm) -- ++(6cm,0);
					\draw[red, densely dashed, line width=1.5pt] (26.3,-3.35cm) -- ++(0,-6.8cm);
					\draw[red, densely dashed, line width=1.5pt] (26.3,-10.15cm) -- ++(-6.1cm,0);
					\draw[->] [red, densely dashed, line width=1.5pt] (20.3,-10.15cm) -- ++(0,-2cm);
					\draw [->] [red, densely dashed, line width=1.5pt] (19.9, -9.6) -- ++(0, -2.8);
				}
				\node[scale =1,draw, thick, inner sep=5pt, rounded corners] (box) {%
					\begin{tikzpicture}
						\matrix[column sep=0.2cm, row sep=0.1cm] {
							\node (V4) {\begin{tikzpicture}[scale=0.4] \figthree \end{tikzpicture} }; \\
							\node (V5) {\begin{tikzpicture}[scale=0.4] \figfive \end{tikzpicture}}; \\
							\node (V6) {\begin{tikzpicture}[scale=0.4] \figseven \end{tikzpicture}}; \\
						};
						\node [right=-0.2cm of V4, yshift=2.25cm, red] {\textbf{(a)}\label{fig:V4}};
						\node [right=0.6cm of V5, yshift=2.25cm] {\textbf{(b)}};
						\node [right=-0.9cm of V6, yshift=2.25cm] {\textbf{(c)}};
						\label{fig:v4_v6}
					\end{tikzpicture}
				};
			\end{tikzpicture}
		}
	\end{subfigure}
	\caption{Hierarchical architectures $V4$, $V5$ and $V6$ implemented in \YOLOvEight.}
	\label{fig:hierarchical_yolo_architectures_V4_V5_V6}
\end{figure*}
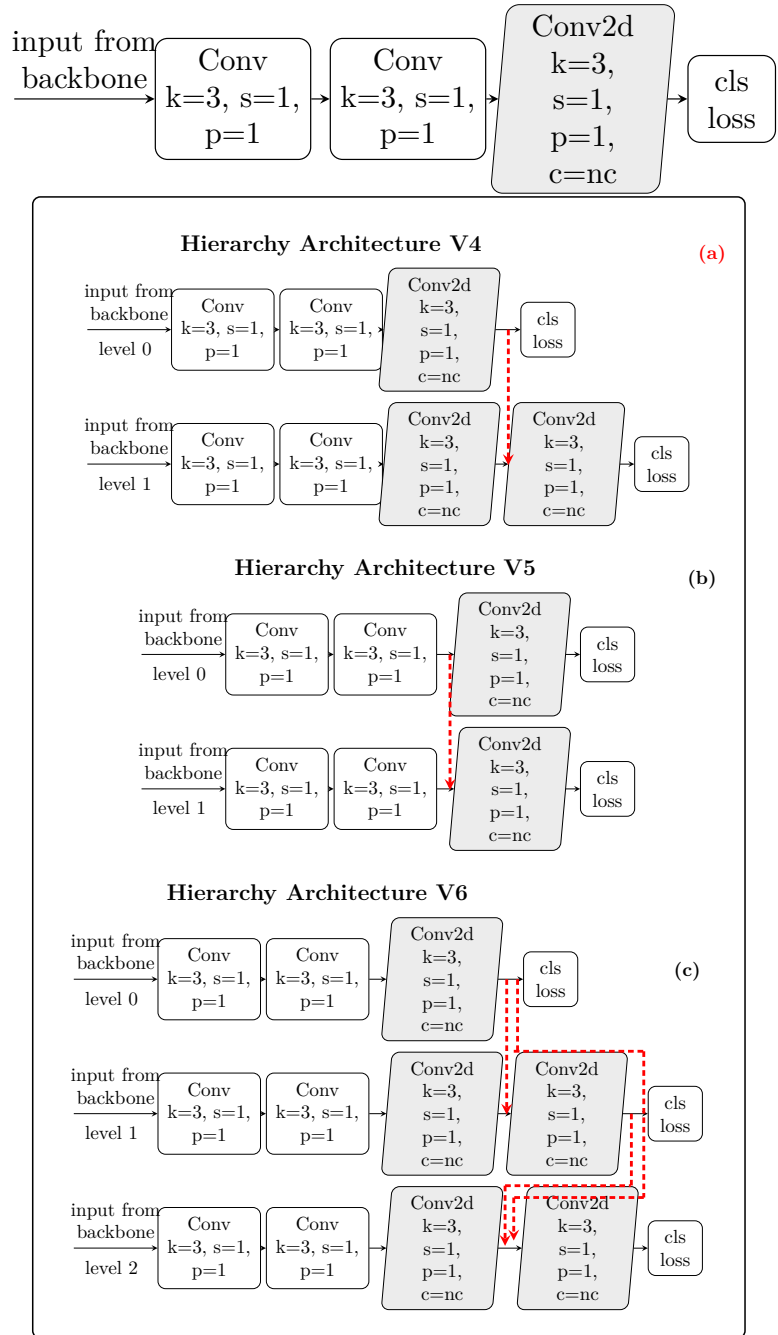

To identify the most effective strategy for integrating hierarchical information, six alternative network architectures were explored. Each design varied in two aspects: the insertion point of the hierarchical branches and the specific location where outputs from preceding levels are concatenated with the current level's features. In all configurations, the output generated by the \YOLOvEight backbone is initially replicated to match the number of hierarchical levels defined in the model. Then, at each level, the output from the previous hierarchical level is concatenated with the current level at a specific point. The dashed red arrows in Figure \ref{fig:hierarchical_yolo_architectures_V1_V2_V3} and Figure \ref{fig:hierarchical_yolo_architectures_V4_V5_V6} indicate these hierarchical connections between levels. The starting point of each arrow represents the layer where the additional hierarchical branch is introduced, while the endpoint signifies the location at the current level where information from the preceding level is concatenated with the current layer’s features.  The different architectural designs were guided by the principle that performance gains in hierarchical models depend on the point at which different levels of abstraction are merged. If lower-level features (e.g., spatial or edge-based) are combined too late in the network, the model may lose valuable spatial context. Conversely, fusing these features too early can overload the classifier with low-level noise, thereby obstructing the learning of high-level semantics (\citet{DBLP:conf/eccv/ZhangZPXS18}). 

In \textit{Version 1} (Figure \ref{fig:hierarchical_yolo_architectures_V1_V2_V3} a), the hierarchical structure is introduced after the Conv2d layer at \textit{level 0}, following a series of convolutional operations through which the model has already extracted low-level features and begun to learn meaningful representations of the input data. However, the concatenation with the subsequent hierarchical \textit{level 1} occurs too early (directly after the backbone output), potentially disrupting the development of higher-order feature representations by introducing prematurely merged signals. This premature fusion may contribute to the notably lower average $F1_{Hier}$ observed for this variant, as reported in Table \ref{tab:architecture_set_metrics}, suggesting that the timing of integration plays a critical role in effective hierarchical learning. 

In the evaluation analysis presented in Table~\ref{tab:architecture_set_metrics}, we compare the six hierarchical model variants (\textit{Version 1} through \textit{6}) using the hierarchical $F1$ score ($F1_{Hier}$) at each level of the hierarchy. A consistent decline in $F1_{Hier}$ is observed from \textit{level 0} to \textit{level 3} across all models, reflecting the increasing complexity and difficulty of fine-grained classification. To identify the most effective architecture, we use two criteria: the $F1_{Hier}$ score at \textit{level 3} which includes the complete set of classes and represents the most granular level of classification, and the $F1_{Hier}$ score of the worst-performing class at this level to capture edge-case performance. This dual-criteria approach enables a more robust assessment of each architectural variant’s capacity to handle the most difficult aspects of the classification task.

\begin{table}[ht]
	\centering
	\caption{{Performance evaluation of various architectural versions based on $F1_{Hier}$ metric for the $100$-class $V1$ dataset.}}
	\label{tab:architecture_set_metrics}
	\renewcommand{\arraystretch}{1.3}
	\begin{tabular*}{\textwidth}{|@{\extracolsep{\fill}} c | c | c | c | c | c |}
		\hline
		\textbf{Arch. Version} & \textbf{Hier Level} & \textbf{Prec\textsubscript{Hier}} & \textbf{Rec\textsubscript{Hier}} & \textbf{ Avg. F1\textsubscript{Hier}} & \textbf{F1\textsubscript{Hier} Worst Class} \\
		\hline
		& 0 & 0.9905 & 0.9954 & 0.9929 & \\
		1 & 1 & 0.9828 & 0.9832 & 0.9830 & \\
		& 2 & 0.9788 & 0.9742 & 0.9765 & \\
		& 3 & 0.9677 & 0.9584 & \textbf{0.9630} & 0.779 \\
		\hline
		& 0 & 0.9951 & 0.9965 & 0.9958 & \\
		2 & 1 & 0.9839 & 0.9822 & 0.9831 & \\
		& 2 & 0.9800 & 0.9731 & 0.9765 & \\
		& 3 & 0.9760 & 0.9543 & \textbf{0.9650} & 0.738 \\
		\hline
		& 0 & 0.9947 & 0.9970 & 0.9958 & \\
		3 & 1 & 0.9878 & 0.9873 & 0.9876 & \\
		& 2 & 0.9828 & 0.9782 & 0.9805 & \\
		& 3 & 0.9730 & 0.9670 & \textbf{0.9700} & 0.800 \\
		\hline
		& 0 & 0.9961 & 0.9971 & 0.9966 & \\
		\textbf{4} & 1 & 0.9890 & 0.9876 & 0.9883 & \\
		& 2 & 0.9833 & 0.9774 & 0.9804 & \\
		& 3 & 0.9734 & 0.9686 & \textbf{0.9710} & \textbf{0.814} \\
		\hline
		& 0 & 0.9957 & 0.9966 & 0.9962 & \\
		5 & 1 & 0.9855 & 0.9843 & 0.9849 & \\
		& 2 & 0.9821 & 0.9764 & 0.9793 & \\
		& 3 & 0.9764 & 0.9598 & \textbf{0.9680} & 0.744 \\
		\hline
		& 0 & 0.9892 & 0.9673 & 0.9781 & \\
		6 & 1 & 0.9887 & 0.9894 & 0.9891 & \\
		& 2 & 0.9838 & 0.9822 & 0.9830 & \\
		& 3 & 0.9741 & 0.9620 & \textbf{0.9680} & 0.701 \\
		\hline
	\end{tabular*}
\end{table}

Building on the insights gained from \textit{Version 1}, modifications aimed at improving the timing of feature integration were explored. In \textit{Version 2} (Figure \ref{fig:hierarchical_yolo_architectures_V1_V2_V3} b), the same insertion point for the hierarchical branch—after the Conv2d layer at \textit{level 0} was kept, but the outputs were concatenated at a later stage: before the Conv2d layer at \textit{level 1}. This adjustment allows the model to learn a richer set of representations at \textit{level 1} prior to the merge. Despite achieving a slightly higher overall $F1_{Hier}$ of $0.965$, this version exhibited a lower worst-class $F1_{Hier}$ of $0.738$. This suggests that  while the delayed concatenation in \textit{Version 2} may slightly enhance global classification performance, it might also suppress critical features for difficult classes, possibly due to delayed access to complementary low-level features.

While \textit{Version 1} suffered from overly premature concatenation and \textit{Version 2} attempted to fix this by delaying fusion, both designs retained the same initial insertion point, potentially limiting their capacity to optimally exploit low-level spatial features. To further investigate the impact of feature timing and semantic alignment, our next architectural variant \textit{Version 3} (Figure \ref{fig:hierarchical_yolo_architectures_V1_V2_V3} c) was designed by shifting both the insertion point and the concatenation point to earlier stages within the network. The idea behind this modification was to allow each hierarchical level to begin processing contextualized features sooner and in closer proximity to the point of integration. By injecting and fusing information earlier in the feature extraction process, the model is given the opportunity to jointly learn spatial and semantic representations in a coordinated fashion. As shown in Table~\ref{tab:architecture_set_metrics}, \textit{Version 3} demonstrated an improvement in both average $F1_{Hier}$ and worst-class performance relative to prior configurations.
\textit{Version 5} (Figure \ref{fig:hierarchical_yolo_architectures_V4_V5_V6} a) retained the same insertion and concatenation scheme employed in \textit{Version 3}, but both points were shifted one convolutional layer deeper into the network. Despite this adjustment, the results showed a decline in both the overall $F1_{Hier}$ and the worst-class $F1_{Hier}$, suggesting that the later integration may have blocked the model's ability to maintain discriminative feature representations at finer levels of classification.

In \textit{Versions 4} and \textit{6}, a second Conv2D layer was introduced at \textit{level 1} and at each subsequent hierarchical level. 
This extra layer serves as a refinement block, ensuring that hierarchical signals are not merely passed along but are processed to resolve conflicting or misaligned representations, and filter out noise or redundancy that may arise from premature or improperly aligned fusion.

Furthermore, in \textit{Version 6}, multiple concatenation points were introduced, specifically from \textit{level 0} to both \textit{levels 1} and \textit{2}. This design choice aimed to enrich the feature representations at higher levels by directly incorporating lower-level information. By allowing these skip connections, the network can more effectively integrate multi-scale features, potentially improving robustness against information loss or degradation that may occur during hierarchical processing. However, this approach did not yield the desired improvements. While the overall $F1_{Hier}$ score remained unchanged, the $F1_{Hier}$ score for the worst-performing class actually decreased. 

Further analysis of Precision and Recall across all six hierarchical architectures (Table~\ref{tab:architecture_set_metrics}) revealed only minor variations at each hierarchical level, indicating that all models generally maintained a well-balanced performance. In addition, all models achieved consistently high metrics, particularly at the base hierarchical level, where both Precision and Recall often exceeded $0.99$. However, \textit{Version 6} exhibited a reduction in Recall at \textit{level 0}, indicating a higher rate of missed positive instances despite maintaining strong Precision. These findings support the selection of \textit{Version 4} as the optimal model, as it provides a more balanced and robust classification performance across all hierarchical levels and achieves the highest overall and worst-performing class $F1_{Hier}$ scores.

\subsection{Datasets}
The complete dataset consisted of $78852$ images of $1011$ grocery store items (classes) and was partitioned into three subsets: a training set of $42456$ images, a validation set of $18198$ images, and a testing set of $18198$ images. The image augmentation process was conducted as follows: for each exposition of a given grocery store item, $17$ random expositions of other items were selected. These items were added to the image sequentially, starting from the larger items and proceeding to the smaller ones. Each item was included only if the maximum allowed occlusion of $70\%$ was not exceeded. Consequently, each image contained approximately $18$ objects on average (Figure \ref{fig:example_image} in the Appendix). 

In addition, the dataset is balanced: each class is represented by approximately $300$ instances in both the validation and testing subsets, and by approximately $700$ instances in the training subset. The instances were generated through permutations of the $10$ different expositions captured by a camera for each item (Figure \ref{fig:segments} in the Appendix). 

Furthermore, to conserve time and computational resources, a reduced dataset comprising only $100$ classes was utilized. This subset was specifically sampled to be statistically representative of the complete dataset and was used in certain experiments to expedite the training process.

\subsection{Dataset Hierarchical Structures}\label{dataset_versions}
Two distinct hierarchical structures of the dataset were developed to evaluate the impact of hierarchical depth and the significance of visual similarities between objects on model performance.

\begin{sidewaysfigure}[htbp!]
	\centering
	\begin{minipage}{0.95\textheight}
		\centering
		\fbox{
			\begin{tikzpicture}[node distance=3cm, scale=0.7, every node/.style={align=center}]
				\node (root) {Root}
				child[xshift=-4.5cm] { node[anchor=center] (food) {\small Food}
					child[xshift=-2cm] { node (canned) {\small Canned \small Goods}
						child { node (meat) {\small Meat} }
						child { node {\small Fish} }
					}
					child[xshift=0cm] { node (dairy) {\small Dairy}
						child { node {\small Cheese} } 
						child { node {Milk} }
					}
					child[xshift=2cm] { node (beverages) {\small Beverages}
						child { node {\small Soda} }
						child { node {\small Juices}
							child[xshift=-1.5cm] { node (apple) {\small Apple \small Juice \small 1.5l. \\ \small Brand A} }
							child[xshift=1.5cm] { node (orange) {\small Orange \small Juice \small 2l. \\ \small Brand B} }
						}
					}
				}
				child[xshift=4.5cm] { node[anchor=center] (nonfood) {\small Non-Food}
					child[xshift=-2cm] { node (house)  {\small Household \\ \small Essentials}
						child[xshift=-0.3cm] { node {Cleaners} }
						child { node {\small Laundry} }
					}
					child[xshift=0cm] { node (health)  {\small Health \\ \& \small Beauty}
						child { node {\small Shampoos} }
						child[xshift=0.5cm] { node {\small Creams} }
					}
					child[xshift=2cm] { node (home) {\small Home \\ \small Goods}
						child { node {\small Kitchen} }
						child { node (storage) {\small Storage} }
					}
				};
				\node[fit=(food) (nonfood), draw, inner sep=0.2cm, yscale=0.7, xscale=1.5, xshift=-0.47cm, rounded corners, lightgray, densely dotted, line width=1pt, label={[anchor=center, yshift=-0.05cm, xshift=-0.9cm]west:\textit{level 0}}] {};
				\node[fit={(canned) (home)}, draw, inner sep=0.2cm, yscale=0.7, xscale=0.98, xshift=-0.13cm, yshift=-0.1cm, rounded corners, lightgray, densely dotted, line width=1pt, label={[anchor=center, yshift=-0.06cm, xshift=-0.9cm]west:\textit{level 1}}] {};
				\node[fit={(meat) (storage)}, draw, inner sep=0.2cm, yscale=0.7, xscale=0.99, xshift=-0.2cm, rounded corners, lightgray, dotted, line width=1pt, label={[anchor=center, yshift=-0.1cm, xshift=-0.9cm]west:\textit{level 2}}] {};
				\node[fit={(apple) (orange)}, draw, inner sep=0.2cm, yscale=0.7, xscale=2.2, , xshift=-0.1cm, rounded corners, lightgray, dotted, line width=1pt, label={[anchor=center, yshift=-0.1cm, xshift=-0.9cm]west:\textit{level 3}}] {};
				
			\end{tikzpicture}
		}
		\captionof{figure}{Dataset Version 1: Categorization with four hierarchical levels, (only a subset of categories shown).}
		\label{fig:dataset_ver_1}
		
		\vspace{2cm}
		\fbox{
			\begin{tikzpicture}[node distance=2cm, scale=0.7, every node/.style={align=center}]
				\node {\small Root}
				child[xshift=-4.5cm] { node[anchor=center] (bags) {\small Bags}
					child[xshift=-2.5cm] { node (20g){\small 20g.}
						child { node (saltines) {\small Saltines} }
						child { node {\small Pretzels} }
					}
					child[xshift=0.5cm] { node {\small 550g.}
						child { node {\small Cereal}
							child[xshift=-1.5cm] { node (brand_a) {\small Brand \small A} child { node (corn_flakes) {\small Corn \\ \small Flakes}}
								child { node {\small Corn \\ \small Puffs}}
							}
							child[xshift=1.5cm] { node {\small Brand \small B} child { node {\small Rice \\ \small Puffs}}
								child { node {\small Rolled \\ \small Oats}}
							}
						}
						child { node {\small Rice} }
					}
				}
				child[xshift=0cm] { node[anchor=center] {\small Bottles}
					child[xshift=-1cm] { node {\small 150ml.}
						child { node {\small Wine} }
						child { node {\small Syrup} }
					}
					child[xshift=1cm] { node {\small 1.5l.}
						child { node {\small Milk} }
						child { node {\small Water} }
					}
				}
				child[xshift=4.5cm] { node[anchor=center] (cans) {\small Cans}
					child[xshift=-0.5cm] { node {\small 250g.}
						child[xshift=0.25cm] { node {\small Beans}
							child[xshift=-1.5cm] { node {\small Brand \small C} child { node {\small Baked \\ \small Beans}}
								child { node {\small Pinto \\ \small Beans}}
							}
							child[xshift=1.5cm] { node (brand_d) {\small Brand \small D} child{ node {\small Black \\ \small Beans}}
								child{ node (mung_beans) {\small Mung \\ \small Beans}}
							}
						}
						child[xshift=0.5cm] { node {\small Corn} }
					}
					child[xshift=1.5cm] { node (500g){\small 500g.}
						child { node {\small Fruit} }
						child { node (puree) {\small Puree} }
					}
				};
				\node[fit=(bags) (cans), draw, inner sep=0.2cm, yscale=0.7, xscale=1.5, xshift=-0.47cm, rounded corners, lightgray, densely dotted, line width=1pt, label={[anchor=center, yshift=-0.05cm, xshift=-0.9cm]west:\textit{level 0}}] {};
				\node[fit={(20g) (500g)}, draw, inner sep=0.2cm, yscale=0.7, xscale=1.1, xshift=-0.13cm, yshift=-0.1cm, rounded corners, lightgray, densely dotted, line width=1pt, label={[anchor=center, yshift=-0.06cm, xshift=-0.9cm]west:\textit{level 1}}] {};
				\node[fit={(saltines) (puree)}, draw, inner sep=0.2cm, yscale=0.7, xscale=0.99, xshift=-0.2cm, rounded corners, lightgray, dotted, line width=1pt, label={[anchor=center, yshift=-0.1cm, xshift=-0.9cm]west:\textit{level 2}}] {};
				\node[fit={(brand_a) (brand_d)}, draw, inner sep=0.2cm, yscale=0.7, xscale=1.25, , xshift=-0.1cm, rounded corners, lightgray, dotted, line width=1pt, label={[anchor=center, yshift=-0.1cm, xshift=-0.9cm]west:\textit{level 3}}] {};
				\node[fit={(corn_flakes) (mung_beans)}, draw, inner sep=0.2cm, yscale=0.7, xscale=1.17, xshift=-0.1cm, rounded corners, lightgray, dotted, line width=1pt, label={[anchor=center, yshift=-0.1cm, xshift=-0.9cm]west:\textit{level 4}}] {};
				
			\end{tikzpicture}
		}
		\captionof{figure}{Dataset Version 2: Categorization with five hierarchical levels.}
		\label{fig:dataset_ver_2}
	\end{minipage}
\end{sidewaysfigure}

\subsubsection{Dataset Hierarchy Version 1:}
In this hierarchical structure the class relations were solely based on store categories, disregarding visual resemblances between objects. This decision was made for the sake of convenience, ensuring that the hierarchy followed a typical supermarket categorization system, starting with broad categories and gradually refining towards more detailed classifications of items within the inventory. The classes were distributed across four hierarchical levels as shown in Figure \ref{fig:dataset_ver_1}. For instance, at \textit{level 0}, classes were divided into food and non-food categories. At \textit{level 1}, the food category was further divided into canned food, dairy, beverages, etc., while the non-food category was divided into household, beauty, home products, etc. Subsequently, at \textit{level 2}, dairy was further subdivided into milk, cheese, yogurt and other specific categories. At the last level, the classes corresponded precisely to the store inventory system nomenclature. This methodology required no additional mapping between hierarchical model categories and inventory categories, as it directly followed the pre-existing inventory framework.

This hierarchical concept was tested only on the reduced $100$-class dataset, since it did not show promise and underperformed compared to the flat \YOLOvEight model. Despite the structured approach, the hierarchical model failed to achieve the desired accuracy, prompting further investigation into alternative hierarchical strategies. This led to the exploration of a hierarchy based on visual similarities between objects.

\subsubsection{Dataset Hierarchy Version 2:}
In dataset hierarchy version $2$ the hierarchical depth was increased to five levels. In addition, visual similarities between items were considered, particularly at lower hierarchical levels. The shape of an object was deemed to be more informative than its size for this classification task due to significant variations in perceived size caused by changes in camera angle, distance, and zoom. Conversely, the shape typically remained consistent, making it a more reliable feature for classification purposes. Therefore, shape was chosen to be the distinctive feature at the lowest \textit{level 0}, followed by size, type, and brand. For instance, categories at \textit{level 0} which represented the shape of the object, included bag, bottle, box, can, cylinder, etc., while \textit{level 1} indicating the size of the object comprised of categories like $1000g.$, $1000ml.$, $100g.$, $10g.$, etc. Higher levels reflected the store's nomenclature: for example, \textit{level 3} described the type of the product such as toothbrush, water, flour, while \textit{level 4} indicated the manufacturer's name, or brand. Once again, classes at \textit{level 4} precisely matched the store's inventory system nomenclature (Figure \ref{fig:dataset_ver_2}).

\subsection{\textbf{\YOLO} \xspace Model and Hyperparameters}
The $V4$ architecture explained in Sec. \ref{hier_architectures}, and the modified loss function defined with Equation \ref{eq:modified_loss2} were both implemented into \YOLOvEight. Pre-trained weights from the COCO dataset were utilized not only to achieve faster convergence and improve model generalization, but also to ensure comparability between models with different architectures. Thus, a standardized starting point for all models was provided, enabling fair comparison across distinct architectures. 

While the model hyperparameters remained at their default settings, the classification loss weight parameter $w_{\text{cls}}$ was modified to better align the training objective with our task. This modification reflects the understanding that different applications may require varying emphasis on classification versus localization. In the \YOLO \xspace framework: the total loss comprises three components: the Complete IoU (CIoU) loss, which serves a bounding box regression loss, the Distribution Focal Loss (DFL), which refines the localization precision of predicted boxes, and the Binary Cross-Entropy (BCE) loss, which is used for classification. Each component contributes to the final loss according to its designated weight: $7.5$ for CIoU (bounding box), $1.5$ for DFL, and $0.5$ for BCE (classification). These weights reflect the relative importance assigned to localization, objectness confidence, and class prediction during training. The choice of these weights, however, should be task-dependent. In applications focused primarily on object detection, such as autonomous driving, which requires accurate identification of potholes on roadways, the relative importance shifts from classification to localization. In these scenarios, misclassifying a roadside artifact as a pothole is preferable to missing the detection altogether, as the vehicle can still navigate around it. Therefore, increasing the weight of the box loss $w_{\text{box}}$ is justified in detection-focused applications, where localization accuracy should be prioritized over classification loss. Ultimately, the tuning of individual loss weights, including $w_{\text{cls}}$ and the box loss weights $w_{\text{box}}$, should be guided by the specific objectives of the application.

\begin{figure*}[htbp!]
	\centering
	\includegraphics[width=\linewidth]{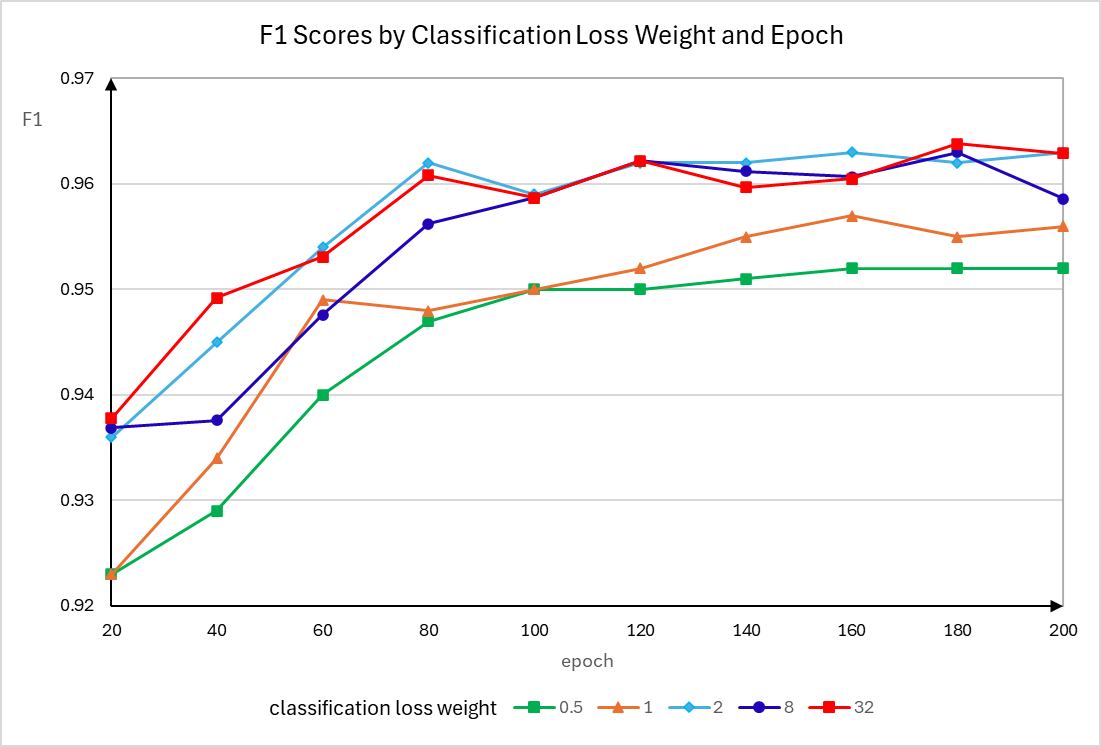}
	\caption{Comparison of $F1$ scores on the test dataset across different classification loss weights ($100$-class dataset).}
	\label{fig:cls_gain}
\end{figure*}

Given that our application focuses on the fine-grained classification of grocery store items, accurate categorization is more important than precise localization. For instance, even if an object is slightly misaligned, shifted to the left or right, it should still be correctly classified. However, misclassifications such as confusing two visually similar but distinct products, can significantly reduce the system’s practical value. To better reflect this priority in the model's training objective, we experimented with increasing the classification loss weight $w_{\text{cls}}$ to assess its impact on the $F1$ score. As shown in Figure \ref{fig:cls_gain}, when training a classification model with the reduced dataset of $100$ classes, the $F1$ score increased with $w_{\text{cls}}$, starting at $0.948$ at $w_{\text{cls}}=0.5$, ultimately reaching its peak of $0.962$ at $w_{\text{cls}}=2$. Therefore, the default value of $w_{\text{cls}} = 0.5$ was changed to $2$. In addition, SGD optimizer was favored over Adam since it demonstrated superior stability during training for our particular model and dataset (refer to Section~\ref{listings} in the Appendix for a comprehensive overview of the configuration parameters).

\subsection{Model Training and Evaluation}

Hierarchy Architecture $V4$ was chosen over the alternatives due to its highest scores across all classes as well as in the worst performing class. 

Subsequently, the model was trained on the two hierarchical versions of the dataset, described in Sec. \ref{dataset_versions}, using the same $V4$ architecture for both. To assess the model's performance, three key metrics were considered: the $F1$ score, True Positive (TP) confidence score, and False Positive (FP) confidence score. The TP confidence score for a specific class is defined by averaging the confidence scores across all true positive predictions associated with that class. Similarly, the FP confidence score is calculated by averaging the confidence scores of all false positive predictions for that specific class. Finally, the TP and FP confidence scores for the entire model are computed as the averages of the TP, or FP scores, respectively, across all classes.

Dataset Hierarchy Version 2, which accounted for the visual and semantic similarities between objects, demonstrated superior performance when applied to the $100$-class dataset. This version outperformed both its predecessor, Version 1, and the flat \YOLOvEight model across all three performance metrics: precision, recall, and F1 score (Table \ref{tab:V1_vs_V2_dataset}). In particular, it achieved the highest TP confidence scores, indicating a greater certainty in its correct predictions. Additionally, it recorded the lowest FP confidence score, suggesting that its incorrect predictions were made with relatively lower confidence, thus enhancing its reliability.

To ensure a fair comparison with the flat \YOLOvEight model, which does not naturally support hierarchical classification, five separate flat models were trained, each corresponding to a distinct hierarchical level within the Dataset Version 2 hierarchy. Moreover, because the hierarchical version of the $F1$ metric is not applicable to flat models, standard precision, recall, and $F1$ scores were used for all evaluations to maintain consistency across comparisons.

\begin{table}[htbp!]
	\centering
	\caption{Comparison between V1, V2 dataset Hierarchies of hYOLO and Flat \YOLOvEight Model using the $100$-class dataset on $V4$ Architecture. Note that the Flat \YOLOvEight model is evaluated using the V2 dataset.}
	\label{tab:V1_vs_V2_dataset}
	\renewcommand{\arraystretch}{1.5}
		\begin{tabular}{|l|c|c|c|c|c|c|}
			\hline
			\textbf{Model Type} & \textbf{Hier. Level} & \textbf{Precision} & \textbf{Recall} & \textbf{F1 Score} & \textbf{TP Conf} & \textbf{FP Conf} \\
			\hline
			& 0 & 0.9961 & 0.9961 & 0.9961 & 0.9555 & 0.1295 \\
			Dataset Version 1 & 1 & 0.9882 & 0.9830 & 0.9856 & 0.9304 & 0.0978 \\
			& 2 & 0.9783 & 0.9721 & 0.9751 & 0.9326 & 0.0705 \\
			& 3 & 0.9552 & 0.9564 & 0.9558 & 0.9179 & 0.0879 \\
			\hline
			& 0 & 0.9922 & 0.9941 & 0.9932 & 0.9592 & 0.0833 \\
			& 1 & 0.9778 & 0.9784 & 0.9781 & 0.9427 & 0.0855 \\
			Dataset Version 2 & 2 & 0.9790 & 0.9771 & 0.9781 & 0.9457 & 0.0679 \\
			& 3 & 0.9748 & 0.9777 & 0.9763 & 0.9448 & 0.0683 \\
			& 4 & 0.9660 & 0.9648 & \textit{\textbf{0.9654}} & \textit{\textbf{0.9406}} & \textit{\textbf{0.0765}} \\
			\hline
			& 0 & 0.9899 & 0.9919 & 0.9909 & 0.9488 & 0.1548 \\
			& 1 & 0.9711 & 0.9745 & 0.9728 & 0.9354 & 0.1259 \\
			Flat \YOLOvEight      & 2 & 0.9835 & 0.9768 & 0.9801 & 0.9374 & 0.1002 \\
			& 3 & 0.9741 & 0.9722 & 0.9731 & 0.9332 & 0.4123 \\
			& 4 & 0.9659 & 0.9610 & 0.9635 & 0.9298 & 0.1253 \\
			\hline
		\end{tabular}
\end{table}

When extended to the more complex $1011$-class dataset, Dataset Hierarchy Version 2 again exhibited lower FP confidence values and higher TP confidence than the flat model (Table \ref{tab:V2_vs_flat_dataset}). We also analyzed the fraction of false positive predictions that fell within the same subgraph as the ground-truth label. This metric reflects semantic closeness even in the presence of prediction errors. Dataset Version 2 significantly outperformed the flat model on this dimension achieving a subgraph-level FP fraction of $0.6844$ at the last hierarchy level 4, compared to $0.6508$ for the flat model.

Interestingly, despite the hierarchical model's superior confidence calibration and semantic consistency, the flat \YOLOvEight model exhibited a higher $F1$ score. However, this comes with the trade-off of increased FP confidence, and a higher likelihood of semantically irrelevant misclassifications, as indicated by the lower fraction of FPs falling within the correct subgraph, underscoring the value of embedding visual similarity into the dataset hierarchy.

\begin{table}[htbp!]
	\centering
	\caption{Comparison between V2 dataset hierarchy of hYOLO and Flat \YOLOvEight Model using the $1011$-class dataset on $V4$ Architecture.}
	\label{tab:V2_vs_flat_dataset}
	\renewcommand{\arraystretch}{1.5}
		\begin{tabular}{|c|c|c|c|c|c|c|c|}
			\hline
			\textbf{Model} & \textbf{HLevel} & \textbf{Prec.} & \textbf{Recall} & \textbf{F1 Score} & \textbf{TP Conf} & \textbf{FP Conf} & \makecell{\textbf{FP Same} \\ \textbf{Subgraph}} \\
			\hline
			& 0 & 0.9914 & 0.9880 & 0.9897 & 0.9423 & 0.0662 &  \\
			Dataset & 1 & 0.9512 & 0.9455 & 0.9484 & 0.9147 & 0.0708 &  \\
			V2 & 2 & 0.9556 & 0.9530 & 0.9543 & 0.9235 & 0.0449 &  \\
			& 3 & 0.9660 & 0.9535 & 0.9597 & 0.9299 & 0.0429 &  \\
			& 4 & 0.8891 & 0.8876 & 0.8883 & \textbf{\textit{0.8735}} & \textbf{\textit{0.0730}} & \textbf{\textit{0.6844}} \\
			
			\hline
			& 0 & 0.9899 & 0.9919 & 0.9909 & 0.9488 & 0.1548 &  \\
			Flat & 1 & 0.9711 & 0.9745 & 0.9728 & 0.9354 & 0.1259 &  \\
			\YOLOvEight & 2 & 0.9835 & 0.9768 & 0.9801 & 0.9374 & 0.1002 &  \\
			& 3 & 0.9741 & 0.9722 & 0.9731 & 0.9332 & 0.1089 &  \\
			& 4 & 0.8976 & 0.8887 &\textbf{\textit{0.8931}} & 0.8710 & 0.0970 & 0.6508 \\
			\hline
		\end{tabular}
\end{table}

In subsequent experiments, the loss function was modified by adding a penalty term as defined in Equation \ref{eq:modified_loss2}. Various values of $\alpha$ were tested to evaluate their impact on performance metrics (Table \ref{tab:dep_loss}). While an $\alpha$ value of $50$ may initially appear excessively high, it is, in fact, appropriate since, in some cases, the introduced penalty term can be smaller by an order of magnitude or more compared to the Binary Cross Entropy (BCE) loss. Without appropriately scaling its weight, the penalty's influence on the overall loss function would be negligible.

\begin{table}[!htbp]
	\centering
	\caption{Performance evaluation of hYOLO for different $\alpha$ values on the $100$-class dataset.}
	\label{tab:dep_loss}
	\renewcommand{\arraystretch}{1.5}
	\begin{tabular*}{0.75\textwidth}{|@{\extracolsep{\fill}}>{\centering\arraybackslash}p{1.5cm}|>{\centering\arraybackslash}p{1cm}|>{\centering\arraybackslash}p{1cm}|>{\centering\arraybackslash}p{1cm}|>{\centering\arraybackslash}p{1.2cm}|>{\centering\arraybackslash}p{1.5cm}|}
		\hline
		\textbf{Alpha} & \textbf{Level} & \textbf{TP conf} & \textbf{FP conf} & \textbf{Avg. F1 Score} & \textbf{F1 Worst Class} \\ \hline
		
		& 0 & 0.9590 & 0.0830 & 0.9930 & 0.9810 \\
		& 1 & 0.9430 & 0.0850 & 0.9774 & 0.8810 \\
		0 & 2 & 0.9460 & 0.0680 & 0.9771 & 0.8810 \\
		& 3 & 0.9478 & 0.0680 & 0.9751 & 0.8770 \\
		& 4 & 0.9410 & \textit{\textbf{0.0760}} & \textit{\textbf{0.9610}} & 0.4850 \\
		\hline
		& 0 & 0.9594 & 0.0919 & 0.9918 & 0.9790 \\
		& 1 & 0.9407 & 0.1191 & 0.9729 & 0.8170 \\
		0.5 & 2 & 0.9462 & 0.0982 & 0.9724 & 0.8240 \\
		& 3 & 0.9478 & 0.0903 & 0.9719 & 0.8210 \\
		& 4 & 0.9406 & 0.1025 & 0.9543 & 0.3620 \\ 
		\hline
		& 0 & 0.9667 & 0.0893 & 0.9919 & 0.9570 \\
		& 1 & 0.9530 & 0.1394 & 0.9704 & 0.8100 \\
		0.9 & 2 & 0.9556 & 0.1028 & 0.9716 & 0.8470 \\
		& 3 & 0.9550 & 0.1366 & 0.9713 & 0.8710 \\
		& 4 & \textit{\textbf{0.9447}} & 0.1360 & 0.9528 & 0.4170 \\
		\hline
		& 0 & 0.9492 & 0.0714 & 0.9899 & 0.9750 \\
		& 1 & 0.9314 & 0.1711 & 0.9731 & 0.8950 \\
		25 & 2 & 0.9382 & 0.1404 & 0.9743 & 0.8800 \\
		& 3 & 0.9396 & 0.1675 & 0.9724 & 0.8480 \\
		& 4 & 0.9312 & 0.1566 & 0.9557 & \textit{\textbf{0.5890}} \\
		\hline
		& 0 & 0.9461 & 0.0539 & 0.9913 & 0.9830 \\
		& 1 & 0.9212 & 0.1879 & 0.9762 & 0.8980 \\
		50 & 2 & 0.9273 & 0.1481 & 0.9749 & 0.8500 \\
		& 3 & 0.9350 & 0.1891 & 0.9740 & 0.8430 \\
		& 4 & 0.9251 & 0.1681 & 0.9566 & 0.4950 \\
		\hline
	\end{tabular*}
\end{table}

As shown in Table \ref{tab:dep_loss}, different values of $\alpha$ optimized different performance metrics. For instance, the best TP confidence score was achieved with $\alpha = 0.9$, whereas the highest $F1$ score and best FP confidence was attained when no penalty term was added. Notably, an increase of $\alpha$ led to an improvement of the $F1$ score; however, the TP score decreased. The lowest per-class $F1$ score increased from $0.362$ ($\alpha = 0.5$) to $0.589$ for $\alpha = 25$.

Table \ref{tab:dep_loss} may initially suggest that the "no penalty" setting ($\alpha=0$) outperforms penalized variants due to its highest average $F1$ score. However, it also yields a markedly lower $F1$ score on the worst-performing class ($0.485$), indicating uneven class-wise performance. In contrast, moderate penalty values (e.g., $\alpha=25$) result in slightly lower average $F1$ scores but offer significantly improved robustness, particularly for difficult classes—evidenced by the highest worst-class $F1$ score ($0.589$). These findings suggest that the magnitude of the penalty term needs tuning, as different values of $\alpha$ yield varying trade-offs across key performance metrics. Therefore, selecting an optimal $\alpha$ should be guided by the specific requirements of the task, such as the importance of overall average performance versus robustness to rare classes.

\section{Conclusion}
This paper introduces hYOLO, a hierarchical end-to-end model for image classification tasks, built upon \YOLOvEight. Our approach incorporates a novel hierarchical architecture, modified loss function, and hierarchical labels. We trained and evaluated the model on two hierarchies derived from the same dataset: one based on formal object categorization without considering visual similarities, and another that accounted for visual resemblances between classes. This dual approach allowed us to assess the impact of hierarchical organization on model performance. Our results demonstrate that incorporating a hierarchical structure aware of the visual similarities between classes significantly improves model performance in image detection and classification tasks. This methodology presents itself as a robust framework for future research and application in various domains.

The practical potential of the proposed hYOLO model extends across a wide spectrum of real-world computer vision applications where visual classes follow an inherent hierarchical structure. In \textbf{medical imaging}, hierarchical classification enables more nuanced differentiation between visually similar disease subtypes, such as various forms of tumors, supporting more accurate diagnostics and assisting decision-support systems in identifying rare or emerging conditions within broader pathological families.  In \textbf{retail environments}, applications such as smart shopping carts, self-checkout systems, and inventory management platforms benefit from hierarchical product classification (e.g., \textit{fruit → citrus → orange}) (\cite{DBLP:journals/cin/WeiTXKS20}). This structure improves product recognition under occlusion or varying packaging, while also enhancing robustness to changes in viewpoint, lighting conditions, and partial visibility during real-time use.  In \textbf{autonomous driving}, understanding hierarchies such as \textit{vehicle → truck → delivery truck} or \textit{sign → warning sign → pedestrian crossing} supports more reliable object recognition and context-aware decision-making, contributing to safer navigation and environment-aware behavioral planning. In \textbf{intelligent surveillance and public safety}, hierarchical modeling can improve both precision and interpretability. By recognizing visual similarities within broader semantic groupings—e.g., \textit{tool → hand tool → hammer} versus \textit{weapon → firearm → handgun}, the system can reduce false positives, differentiate harmless from threatening objects, and enhance situational awareness in real time. This is particularly valuable in high-stakes environments such as airports, public venues, or critical infrastructure surveillance, where accurate threat detection is important. In \textbf{e-commerce image classification}, hierarchical understanding may allow platforms to better manage extensive product taxonomies. For instance, classifying a product as \textit{clothing → outerwear → jacket → leather jacket} enables more relevant search results, improved recommendations, and more granular filtering options (\citet{DBLP:journals/eswa/SeoS19}). This will improve user experience and will facilitate more efficient product discovery across vast and diverse online catalogs. In \textbf{ecological and environmental monitoring}, the classification of flora, fauna, and land cover types aligns naturally with taxonomic hierarchies (e.g., \textit{animal → bird → raptor → eagle}), thus supporting scalable biodiversity assessment, species monitoring, and habitat mapping, which is essential for conservation science (\cite{f9a35fdca67d494f8151f43e08964769}). Finally, in \textbf{digital art and cultural heritage preservation}, classifying artifacts by attributes such as \textit{art → painting → Renaissance → Italian} may enhance digital curation, retrieval, and archival efforts in museums and academic institutions (\cite{DBLP:conf/icpr/JainBBMOK20} and \citet{DBLP:conf/qurator/Neudecker22}). This will facilitate more structured metadata creation and will support automated tagging for improved user interaction and historical analysis.

The promising results achieved with our hierarchical model highlight several avenues for future research, including exploring deeper hierarchies, alternative architectures, enhanced loss functions, establishing standardized guidelines for creating hierarchical datasets and best practices for defining hierarchical levels, categories, and labeling conventions, developing tools and metrics for evaluating the quality of hierarchical datasets, and addressing the challenge of class imbalance within hierarchical datasets, particularly at deeper levels of the hierarchy. By following these research paths, further advancement of hierarchical modeling will be achieved, ensuring robust and reliable performance in practical applications.

\section{Data Availability}
The datasets generated and analyzed in this study are not publicly available due to proprietary restrictions. The data constitute a private dataset, created with significant investment in its generation, and are subject to copyright and intellectual property protections. Sharing the complete dataset would violate proprietary considerations. However, we provide a representative subset containing 100 classes, which reflects the diversity and structure of the full dataset and is sufficient to reproduce the results presented in this paper. This datsaset is available upon reasonable request.

\section{Code Availability}
The code to reproduce the experiments described in this paper is available at the following repository: \url{https://github.com/ds2run/hyolo}

\bibliography{sn-bibliography}

\clearpage 
\onecolumn 

\begin{appendices}
	
	\section{Hierarchical Model Training Workflow}\label{workflow}
	
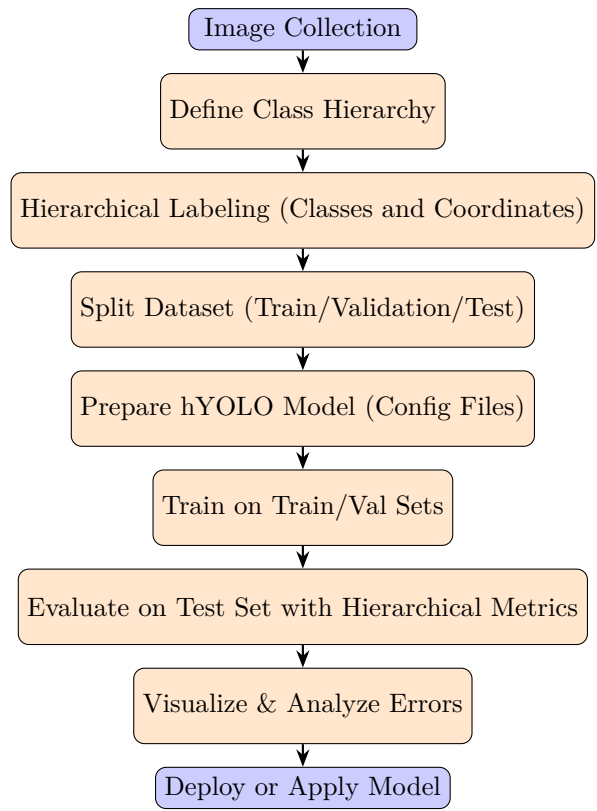
\begin{figure}[H]
		\centering
		\begin{tikzpicture}[node distance=0.3cm]
			\tikzstyle{startstop} = [rectangle, rounded corners, minimum width=3cm, minimum height=0.5cm, text centered, draw=black, fill=blue!20]
			\tikzstyle{process} = [rectangle, rounded corners, minimum width=3.5cm, minimum height=1cm, text centered, draw=black, fill=orange!20]
			\tikzstyle{arrow} = [thick, ->, >=Stealth]
			
			\node (start) [startstop] {Image Collection};
			\node (hierarchy) [process, below=of start] {Define Class Hierarchy};
			\node (label) [process, below=of hierarchy] {Hierarchical Labeling \par (Classes and Coordinates)};
			\node (split) [process, below=of label] {Split Dataset \par (Train/Validation/Test)};
			\node (model) [process, below=of split] {Prepare hYOLO Model \par (Config Files)};
			\node (train) [process, below=of model] {Train on Train/Val Sets};
			\node (evaluate) [process, below=of train] {Evaluate on Test Set with Hierarchical Metrics};
			\node (analyze) [process, below=of evaluate] {Visualize \& Analyze Errors};
			\node (deploy) [startstop, below=of analyze] {Deploy or Apply Model};
			
			\draw [arrow] (start) -- (hierarchy);
			\draw [arrow] (hierarchy) -- (label);
			\draw [arrow] (label) -- (split);
			\draw [arrow] (split) -- (model);
			\draw [arrow] (model) -- (train);
			\draw [arrow] (train) -- (evaluate);
			\draw [arrow] (evaluate) -- (analyze);
			\draw [arrow] (analyze) -- (deploy);
		\end{tikzpicture}
		\caption{Workflow diagram for hierarchical object detection and classification using the hYOLO framework.}
		\label{fig:workflow_hYOLO}
	\end{figure}

	The process to train a hierarchical object detection and classification model involves several well-defined stages. The workflow (Figure \ref{fig:workflow_hYOLO}) begins with data preparation and annotation structured according to a class hierarchy, followed by dataset partitioning and model configuration. Subsequent steps focus on training, evaluation using hierarchical metrics, error analysis, and model deployment for practical applications. The following outlines each step in detail:
	
	\begin{itemize}
		\item \textbf{Image Collection:} This initial step involves gathering a comprehensive set of images relevant to the problem domain. Both the quality and diversity of the collected images directly impact the robustness and generalizability of the model developed later. Additionally, it is important to include images with multiple objects present, potentially overlapping, to better simulate real-world scenarios. To further enhance the dataset, various data augmentation techniques such as rotation, scaling, flipping, and color adjustments should be applied to increase variability and improve the model's ability to generalize across different conditions.
		
		\item \textbf{Define Class Hierarchy:} A meaningful and structured taxonomy of classes has to be established to represent the relationships between different categories, ideally based on visual similarities or semantic relevance. This hierarchical organization helps the model utilize class dependencies effectively during training.
		
		\item \textbf{Hierarchical Labeling (Classes and Coordinates):} In this step, each image is annotated not only with its multi-level class label but also with precise object coordinates such as bounding boxes. Hierarchical labeling ensures that labels respect the defined class structure, enhancing the model’s ability to detect and classify objects at multiple levels.
		
		\item \textbf{Split Dataset (Train/Validation/Test):} The dataset is partitioned into training, validation, and testing subsets. This division is important for unbiased model training, hyperparameter tuning, and performance evaluation on unseen data.
		
		\item \textbf{Prepare hYOLO Model (Config Files, see Section~\ref{listings}):} This step involves configuring the hierarchical YOLO (hYOLO) model to handle multi-level object detection and classification. It typically requires modifying configuration files, such as the default YAML, to specify the hierarchical class structure, including the names and number of classes at each level. In addition, essential parameters set during this step include the number of training epochs, commonly 300, which controls the total iterations over the dataset; the batch size, which determines how many images are processed simultaneously in each training iteration; and the input image size, often set to 1280 pixels, which affects detection precision and computational requirements. Overall, this step ensures that the model is properly configured to effectively learn both object localization and classification across the multiple hierarchical levels represented in the data.
		
		\item \textbf{Train on Train/Val Sets:} The model is trained using the training dataset, with the validation set used to monitor progress and prevent overfitting. Iterative updates are made to the model parameters to minimize the loss function and improve predictive accuracy.
		
		\item \textbf{Evaluate on Test Set with Hierarchical Metrics:} The trained model’s performance is assessed on the test set using hierarchical metrics. This step provides an objective measure of how well the model generalizes to new, unseen data.
		
		\item \textbf{Visualize \& Analyze Errors:} Results and misclassifications are visualized to identify patterns of errors or weaknesses in the model. This analysis helps identify areas for improvement, such as enhancing data augmentation, fine-tuning the model parameters, or adjusting the training strategy.
		
		\item \textbf{Deploy or Apply Model:} Finally, the validated model is deployed in a real-world environment or integrated into an application. This deployment may involve embedding the model on edge devices, such as Raspberry Pi, NVIDIA Jetson Nano, or other compact embedded systems, enabling real-time object classification directly within the operational environment. This step also includes ongoing monitoring of the model’s performance post-deployment and planning for future updates or refinements as necessary.
	\end{itemize}
	
	\section{Training Dataset Samples}\label{train_sample}
	
	\begin{figure}[htb!]
		\centering
		\includegraphics[width=\linewidth]{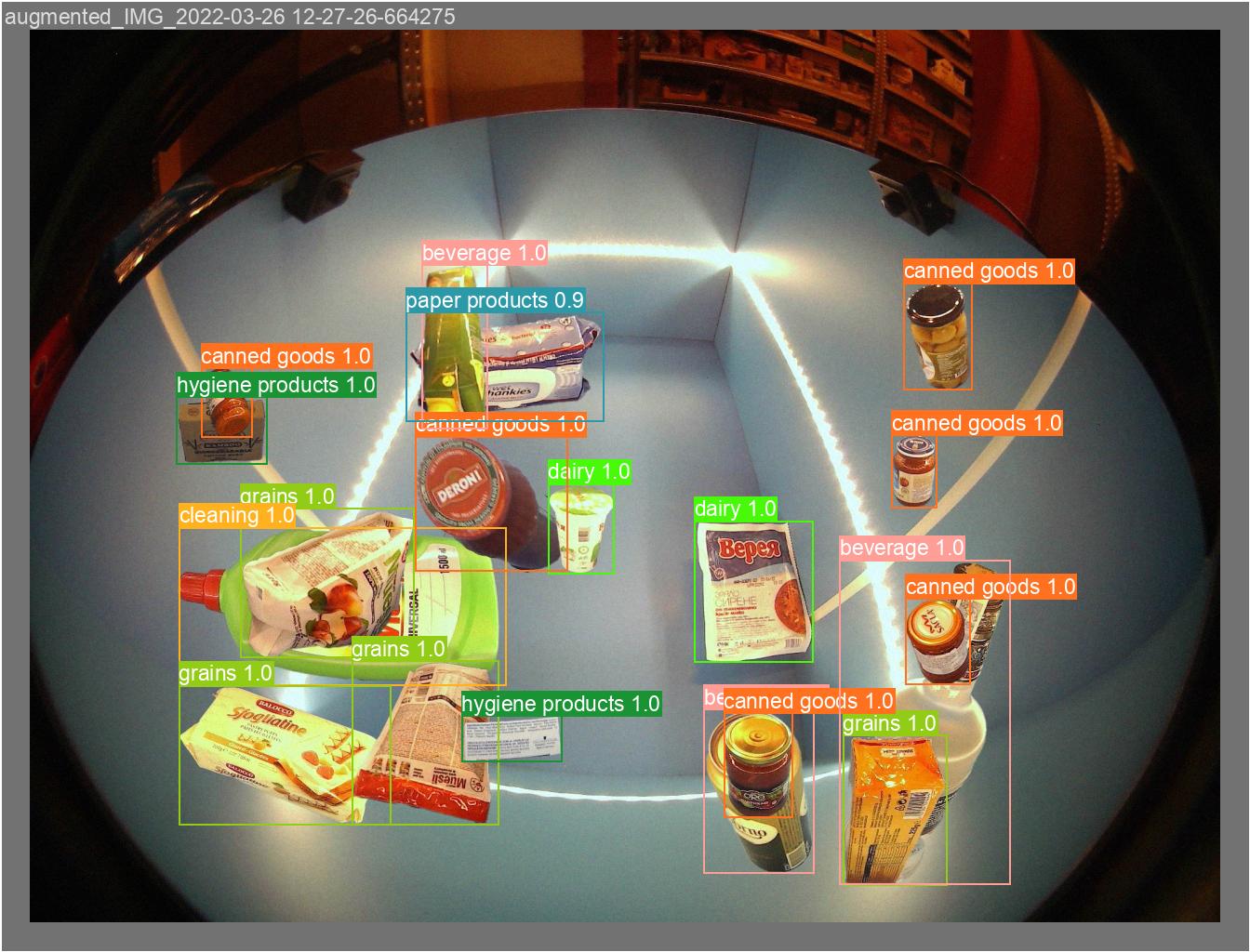}
		\caption{An example image from the dataset.}
		\label{fig:example_image}
	\end{figure}
	
	\begin{figure}[htb!]
		\centering
		\fbox{\rule{0pt}{0.5in} 
		 \includegraphics[width=0.95\linewidth]{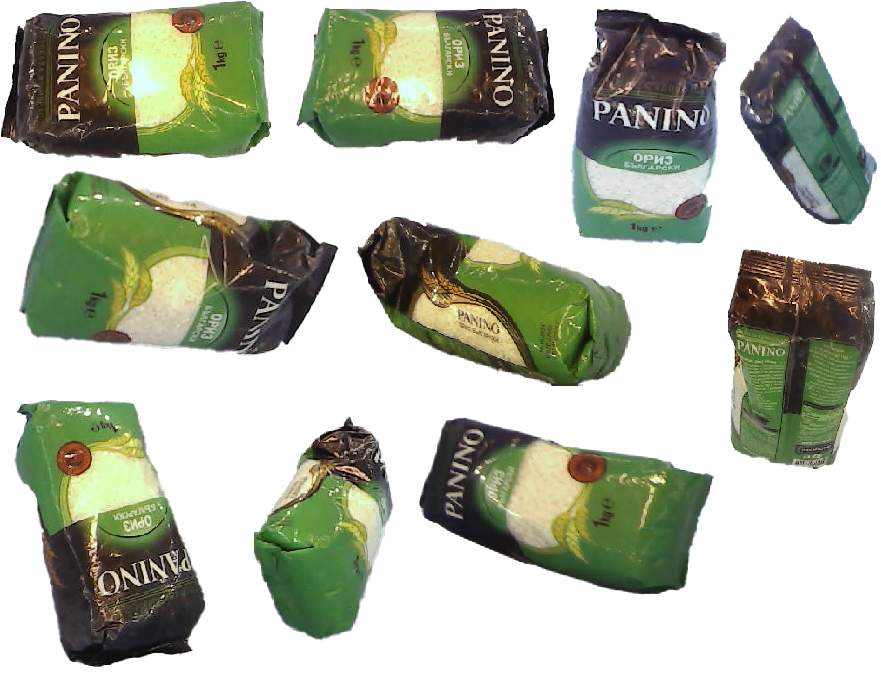}}
	\caption{Different expositions of the same object.}
	\label{fig:segments}
\end{figure}

\clearpage
\section{Listings}\label{listings}

{\scriptsize
	\begin{lstlisting}[language=YAML,label={lst:yaml}, caption={Changes in the default parameters in \YOLOvEight in \href{https://github.com/ultralytics/ultralytics/blob/main/ultralytics/cfg/default.yaml}{cfg/default.yaml}} when adding hierarchy ]
		
		# Ultralytics YOLO, GPL-3.0 license
		# Default training settings and hyperparameters for medium-augmentation COCO training
		
		task: detect  # YOLO task, i.e. detect, segment, classify, pose
		mode: train  # YOLO mode, i.e. train, val, predict, export, track, benchmark
		#hier_yolo: True
		#hierarchy_names:
		hier_depth: 5
		calc_TP_FN_FP: False
		calc_set_metric: False
		get_hier_paths: False
		calc_TP_FP_conf: False
		dependency_loss: True
		
		# Train settings
		-----------------------------------------------------------------------------------
		model:  # path to model file, i.e. yolov8n.pt, yolov8n.yaml
		data:  # path to data file, i.e. coco128.yaml
		epochs: 300  # number of epochs to train for
		patience: 30  # epochs to wait for no observable improvement for early stopping of training
		batch: 1  # number of images per batch (-1 for AutoBatch)
		imgsz: 1280  # size of input images as integer or w,h
		save: True  # save train checkpoints and predict results
		save_period: 10 # Save checkpoint every x epochs (disabled if < 1)
		cache: disk  # True/ram, disk or False. Use cache for data loading
		device: cpu # device to run on, i.e. cuda device=0 or device=0,1,2,3 or device=cpu
		workers: 8  # number of worker threads for data loading (per RANK if DDP)
		project:  runs/hier_yolo # project name
		name:  #train_100class_hier_v2 # experiment name, results saved to 'project/name' directory
		exist_ok: False  # whether to overwrite existing experiment
		pretrained: False  # whether to use a pretrained model
		optimizer: SGD  # optimizer to use, choices=['SGD', 'Adam', 'AdamW', 'RMSProp']
		verbose: True  # whether to print verbose output
		seed: 0  # random seed for reproducibility
		deterministic: True  # whether to enable deterministic mode
		single_cls: False  # train multi-class data as single-class
		image_weights: False  # use weighted image selection for training
		rect: False  # support rectangular training if mode='train', support rectangular evaluation if mode='val'
		cos_lr: False  # use cosine learning rate scheduler
		close_mosaic: 0  # disable mosaic augmentation for final 10 epochs
		resume: False  # resume training from last checkpoint
		amp: False  # Automatic Mixed Precision (AMP) training, choices=[True, False], True runs AMP check
		# Segmentation
		overlap_mask: True  # masks should overlap during training (segment train only)
		mask_ratio: 4  # mask downsample ratio (segment train only)
		# Classification
		dropout: 0.0  # use dropout regularization (classify train only)
		
		# Val/Test settings
		-----------------------------------------------------------------------------------
		val: True  # validate/test during training
		split: val  # dataset split to use for validation, i.e. 'val', 'test' or 'train'
		save_json: False  # save results to JSON file
		save_hybrid: False  # save hybrid version of labels (labels + additional predictions)
		conf:  # object confidence threshold for detection (default 0.25 predict, 0.001 val)
		iou: 0.7  # intersection over union (IoU) threshold for NMS
	\end{lstlisting}
}

{\scriptsize
	\begin{lstlisting}[language=YAML]
		max_det: 300  # maximum number of detections per image
		half: False  # use half precision (FP16)
		dnn: False  # use OpenCV DNN for ONNX inference
		plots: True  # save plots during train/val
		
		# Prediction settings
		-----------------------------------------------------------------------------------
		source:  # source directory for images or videos
		show: False  # show results if possible
		save_txt: False  # save results as .txt file
		save_conf: False  # save results with confidence scores
		save_crop: False  # save cropped images with results
		hide_labels: False  # hide labels
		hide_conf: False  # hide confidence scores
		vid_stride: 1  # video frame-rate stride
		line_thickness: 3  # bounding box thickness (pixels)
		visualize: False  # visualize model features
		augment: False  # apply image augmentation to prediction sources
		agnostic_nms: False  # class-agnostic NMS
		classes:  # filter results by class, i.e. class=0, or class=[0,2,3]
		retina_masks: False  # use high-resolution segmentation masks
		boxes: True  # Show boxes in segmentation predictions
		
		# Export settings
		-----------------------------------------------------------------------------------
		format: torchscript  # format to export to
		keras: False  # use Keras
		optimize: False  # TorchScript: optimize for mobile
		int8: False  # CoreML/TF INT8 quantization
		dynamic: False  # ONNX/TF/TensorRT: dynamic axes
		simplify: False  # ONNX: simplify model
		opset:  # ONNX: opset version (optional)
		workspace: 4  # TensorRT: workspace size (GB)
		nms: False  # CoreML: add NMS
		
		# Hyperparameters
		-----------------------------------------------------------------------------------
		lr0: 0.01  # initial learning rate (i.e. SGD=1E-2, Adam=1E-3)
		lrf: 0.01  # final learning rate (lr0 * lrf)
		momentum: 0.937  # SGD momentum/Adam beta1
		weight_decay: 0.0005  # optimizer weight decay 5e-4
		warmup_epochs: 3.0  # warmup epochs (fractions ok)
		warmup_momentum: 0.8  # warmup initial momentum
		warmup_bias_lr: 0.1  # warmup initial bias lr
		box: 7.5  # box loss gain
		cls: 2.0  # cls loss gain (scale with pixels)
		dfl: 1.5  # dfl loss gain
		fl_gamma: 0.0  # focal loss gamma (efficientDet default gamma=1.5)
		label_smoothing: 0.0  # label smoothing (fraction)
		nbs: 64  # nominal batch size
		hsv_h: 0.015  # image HSV-Hue augmentation (fraction)
		hsv_s: 0.7  # image HSV-Saturation augmentation (fraction)
		hsv_v: 0.4  # image HSV-Value augmentation (fraction)
		degrees: 0.0  # image rotation (+/- deg)
		translate: 0.1  # image translation (+/- fraction)
		scale: 0.5  # image scale (+/- gain)
		shear: 0.0  # image shear (+/- deg)
		perspective: 0.0  # image perspective (+/- fraction), range 0-0.001
		flipud: 0.0  # image flip up-down (probability)
		fliplr: 0.5  # image flip left-right (probability)
		mosaic: 0.0  # image mosaic (probability)
		mixup: 0.0  # image mixup (probability)
		copy_paste: 0.0  # segment copy-paste (probability)
		
		
	\end{lstlisting}
}
\end{appendices}

\end{document}